%% file: main.tex
\documentclass[letterpaper, 10 pt, journal, oneside]{IEEEtran}
\pdfoutput=1

\usepackage[utf8]{inputenc} 
\usepackage{atbegshi,etoolbox}
\usepackage[T1]{fontenc}    
\usepackage{url}            
\usepackage{booktabs}       
\usepackage{amsfonts}       
\usepackage{nicefrac}       
\usepackage{xcolor}         
\usepackage{textcomp, gensymb}
\usepackage{amssymb}
\usepackage{amsmath}
\usepackage{tabularx}
\usepackage{multirow}
\usepackage{multicol}
\usepackage{xspace}
\usepackage[pdftex]{graphicx}
\usepackage[globalcitecopy]{bibunits}
\defaultbibliographystyle{IEEEtran} 
\defaultbibliography{main}
\usepackage[caption=false,font=footnotesize]{subfig}
\usepackage{stackrel}
\usepackage{wrapfig}
\usepackage{endnotes}
\usepackage{siunitx}
\usepackage{enumitem}
\usepackage{hyperref}
\hypersetup{colorlinks=true,breaklinks=true,citecolor=blue,linkcolor=red,urlcolor=blue}

\DeclareMathOperator{\add}{ADD}
\DeclareMathOperator{\rot}{rot}
\DeclareMathOperator{\tra}{tra}

\DeclareMathOperator{\ratioaug}{aug}
\DeclareMathOperator{\rationoaug}{noaug}
\DeclareMathOperator{\ratio}{ratio}
\DeclareMathOperator{\opref}{ref}
\DeclareMathOperator{\oppre}{pre}
\newcommand{\newcmdcolor}[3]{
  \newcommand{#1}{\perhapscolorize{#2}{#3}}%
}

\usepackage{etoolbox}

\makeatletter

\def\hyper@natlinkstart#1{%
  \Hy@backout{#1}%
  \hyper@linkstart{cite}{cite.\@bibunitname.#1}%
  \def\hyper@nat@current{#1}%
}

\def\hyper@natlinkbreak#1#2{%
  \hyper@linkend#1\hyper@linkstart{cite}{cite.\@bibunitname.#2}%
}

\def\hyper@natanchorstart#1{%
  \hyper@anchorstart{cite.\@bibunitname.#1}%
}

\def\bibcite#1#2{%
  \@newl@bel{b}{#1}{\hyper@@link[cite]{}{cite.\@bibunitname.#1}{#2}}%
}%

\def\@lbibitem[#1]#2{%
  \@skiphyperreftrue
  \H@item[\hyper@anchorstart{cite.\@bibunitname.#2}%
  \@BIBLABEL{#1}\hyper@anchorend\hfill]%
  \@skiphyperreffalse
  \if@filesw
    \begingroup
      \let\protect\noexpand
      \immediate\write\@auxout{%
        \string\bibcite{#2}{#1}%
      }%
    \endgroup
  \fi
  \ignorespaces
}%

\def\@bibitem#1{%
  \@skiphyperreftrue\H@item\@skiphyperreffalse
  \hyper@anchorstart{cite.\@bibunitname.#1}\relax\hyper@anchorend
  \if@filesw
    \begingroup
      \let\protect\noexpand
      \immediate\write\@auxout{%
        \string\bibcite{#1}{\the\value{\@listctr}}%
      }%
    \endgroup
  \fi
  \ignorespaces
}%

\def\@citex[#1]#2{%
  \let\@citea\@empty
  \@cite{%
    \@for\@citeb:=#2\do{%
      \@citea
      \def\@citea{,\penalty\@m\ }%
      \edef\@citeb{\expandafter\@firstofone\@citeb}%
      \if@filesw
        \immediate\write\@auxout{\string\citation{\@citeb}}%
      \fi
      \@ifundefined{b@\@citeb}{%
        \mbox{\reset@font\bfseries ?}%
        \G@refundefinedtrue
        \@latex@warning{%
          Citation `\@citeb' on page \thepage \space undefined%
        }%
      }{%
        \hyper@natlinkstart{\@citeb}%
            \hbox{\csname b@\@citeb\endcsname}%
        \hyper@natlinkend%
      }%
    }%
  }{#1}%
}%

\newcommand{\perhapscolorize}[1]{%
  \ifcolorize
    \expandafter\@firstoftwo
  \else
    \expandafter\@secondoftwo
  \fi
  {\textcolor{#1}}%
  {\@firstofone}%
}
\makeatother
\newif\ifcolorize

\newcommand{\eg}{e.g.~}
\newcommand{\ie}{i.e.~}
\newcommand{\sixd}{6D\xspace}
\newcommand{\rgbd}{RGB-D\xspace}
\newcommand{\dataset}{\textit{Imitrob}\xspace}
\newcmdcolor{\train}{blue}{\textit{ImitrobTrain}\xspace}
\newcmdcolor{\test}{blue}{\textit{ImitrobTest}\xspace}
\newcommand{\gluegun}{glue\,gun\xspace}
\newcommand{\groutfloat}{grout\,float\xspace}
\newcommand{\roller}{roller\xspace}
\newcommand{\gluegunII}{glue\,gun\,2\xspace}
\newcommand{\gluegunIII}{glue\,gun\,3\xspace}
\newcommand{\gluegunIIIdtp}{glue\,gun\,4\xspace}
\newcommand{\heatgun}{heat\,gun\xspace}
\newcommand{\powerdrill}{power\,drill\xspace}
\newcommand{\solder}{soldering\,iron\xspace}
\newcommand{\numtools}{9\xspace}
\newcommand{\numtoolstext}{nine\xspace}
\newcommand{\numtasks}{12\xspace}
\newcommand{\numtaskstext}{twelve\xspace}

\newcommand{\bgnoise}{\textit{BgNoise}\xspace}
\newcommand{\bgrandom}{\textit{BgRandom}\xspace}
\newcommand{\bgalternate}{\textit{BgAlternate}\xspace}
\newcommand{\bgblend}{\textit{BgBlend}\xspace}
\newcommand{\noaugmentation}{\textit{NoAug}\xspace}
\newcommand{\maskthresholding}{\textit{MaskThresholding}\xspace}
\newcommand{\maskfba}{\textit{MaskFBA}\xspace}
\newcommand{\noclutter}{\textit{NoClutter}\xspace}
\newcommand{\clutter}{\textit{Clutter}\xspace}
\newcommand{\twocm}{2\,cm\xspace}
\newcommand{\fivecm}{5\,cm\xspace}
\newcommand{\tencm}{10\,cm\xspace}
\newcommand{\allsubjects}{AllToAll\xspace}
\newcommand{\differentsubject}{ThreeToDiff\xspace}
\newcommand{\samesubject}{OneToSame\xspace}
\newcommand{\samecamera}{Same\xspace}
\newcommand{\othercamera}{Other\xspace}
\newcommand{\bothcameras}{Both\xspace}
\newcommand{\samehand}{Same\xspace}
\newcommand{\oppositehand}{Other\xspace}
\newcommand{\bothhands}{Both\xspace}
\newcommand{\discardpages}[1]{
  \xdef\discard@pages{#1}
  \AtBeginShipout{
    \renewcommand*{\do}[1]{
      \ifnum\value{page}=##1\relax%
        \AtBeginShipoutDiscard
        \gdef\do####1{}
      \fi%
    }%
    \expandafter\docsvlist\expandafter{\discard@pages}
  }%
}
\newif\ifkeeppage
\newcommand{\keeppages}[1]{
  \xdef\keep@pages{#1}
  \AtBeginShipout{
    \keeppagefalse%
    \renewcommand*{\do}[1]{
      \ifnum\value{page}=##1\relax%
        \keeppagetrue
        \gdef\do####1{}
      \fi%
    }%
    \expandafter\docsvlist\expandafter{\keep@pages}
    \ifkeeppage\else\AtBeginShipoutDiscard\fi
  }%
}

\usepackage{subfiles} 

\begin{document}

\begin{bibunit}

\subfile{secs/paper}
\putbib
\end{bibunit}



\subfile{secs/supmat}


\bibliographystyle{IEEEtran}
\bibliography{main}

\end{document}

%% file: secs/paper.tex
\title{Imitrob: Imitation Learning Dataset for Training and Evaluating 6D Object  Pose Estimators}

\author{Jiri~Sedlar$^{1,*}$, Karla~Stepanova$^{1,*}$, Radoslav~Skoviera$^{1}$, Jan~K.~Behrens$^{1}$, Matus~Tuna$^{2}$, Gabriela~Sejnova$^{1}$, Josef~Sivic$^{1}$, and~Robert~Babuska$^{1,3}$%
\thanks{
This work was supported by the European Regional Development Fund under the project IMPACT (reg.~no.~CZ.02.1.01/0.0/0.0/15\_003/0000468), European Regional Development Fund under project Robotics for Industry 4.0 (reg.~no.~CZ.02.1.01/0.0/0.0/15\_003/0000470),
EU Horizon Europe Programme under the project AGIMUS (reg.~no.~101070165),
VEGA 1/0796/18, MPO TRIO project num. FV40319, CTU Student Grant Agency (reg.~no.~SGS21/184/OHK3/3T/37), and  by the Czech Science Foundation (project no.~GA21-31000S).}
\thanks{$^{1 }$Czech Institute of Informatics, Robotics and Cybernetics, Czech Technical University in Prague, Czech Republic}
\thanks{$^{2 }$Faculty of Mathematics, Physics and Informatics, Comenius University in Bratislava, Slovakia}
\thanks{$^{3 }$Cognitive Robotics, Faculty of 3mE, Delft University of Technology, The Netherlands}
\thanks{$^{* }$Both authors contributed equally. E-mail: jiri.sedlar@cvut.cz, karla.stepanova@cvut.cz}
\thanks{The dataset is available at \url{http://imitrob.ciirc.cvut.cz/imitrobdataset.php}}
\thanks{The code is available at \url{https://github.com/imitrob/imitrob_dataset_code}}
\thanks{\copyright~2023 IEEE. Personal use of this material is permitted. Permission from IEEE must be obtained for all other uses, in any current or future media, including reprinting/republishing this material for advertising or promotional purposes, creating new collective works, for resale or redistribution to servers or lists, or reuse of any copyrighted component of this work in other works.}
\thanks{\url{https://ieeexplore.ieee.org/document/10077123}}
\thanks{Digital Object Identifier (DOI): \href{https://ieeexplore.ieee.org/document/10077123}{10.1109/LRA.2023.3259735}}}

\maketitle




\begin{abstract}
This paper introduces a dataset for training and evaluating methods for 6D pose estimation of hand-held tools in task demonstrations captured by a standard RGB camera. Despite the significant progress of 6D pose estimation methods, their performance is usually limited for heavily occluded objects, which is a common case in imitation learning, where the object is typically partially occluded by the manipulating hand. Currently, there is a lack of datasets that would enable the development of robust 6D pose estimation methods for these conditions. To overcome this problem, we collect a new dataset (Imitrob) aimed at 6D pose estimation in imitation learning and other applications where a human holds a tool and performs a task. The dataset contains image sequences of \numtoolstext different tools and \numtaskstext manipulation tasks with two camera viewpoints, four human subjects, and left/right hand. Each image is accompanied by an accurate ground truth measurement of the 6D object pose obtained by the HTC Vive motion tracking device. The use of the dataset is demonstrated by training and evaluating a recent 6D object pose estimation method (DOPE) in various setups.
\end{abstract}


\section{Introduction}
\IEEEPARstart{D}{espite} the recent progress~\cite{labbe2020cosypose,tremblay2018deep}, \sixd object pose estimators have rarely been applied to image sequences capturing manipulation of hand-held objects. However, such a set-up has a huge potential in imitation learning scenarios where expert demonstrations are used to teach robots new tasks (e.g.~a human demonstrator manipulating a glue gun to apply glue along specified trajectories). One of the reasons is that there are no datasets and benchmarks that would allow training for such setups. To overcome this problem, we have collected and annotated a real-world hand-held tool manipulation dataset (\dataset) that allows training and evaluating \sixd object pose estimators in such conditions. 
\begin{figure}
    \centering
    \includegraphics[width = 0.45\textwidth]{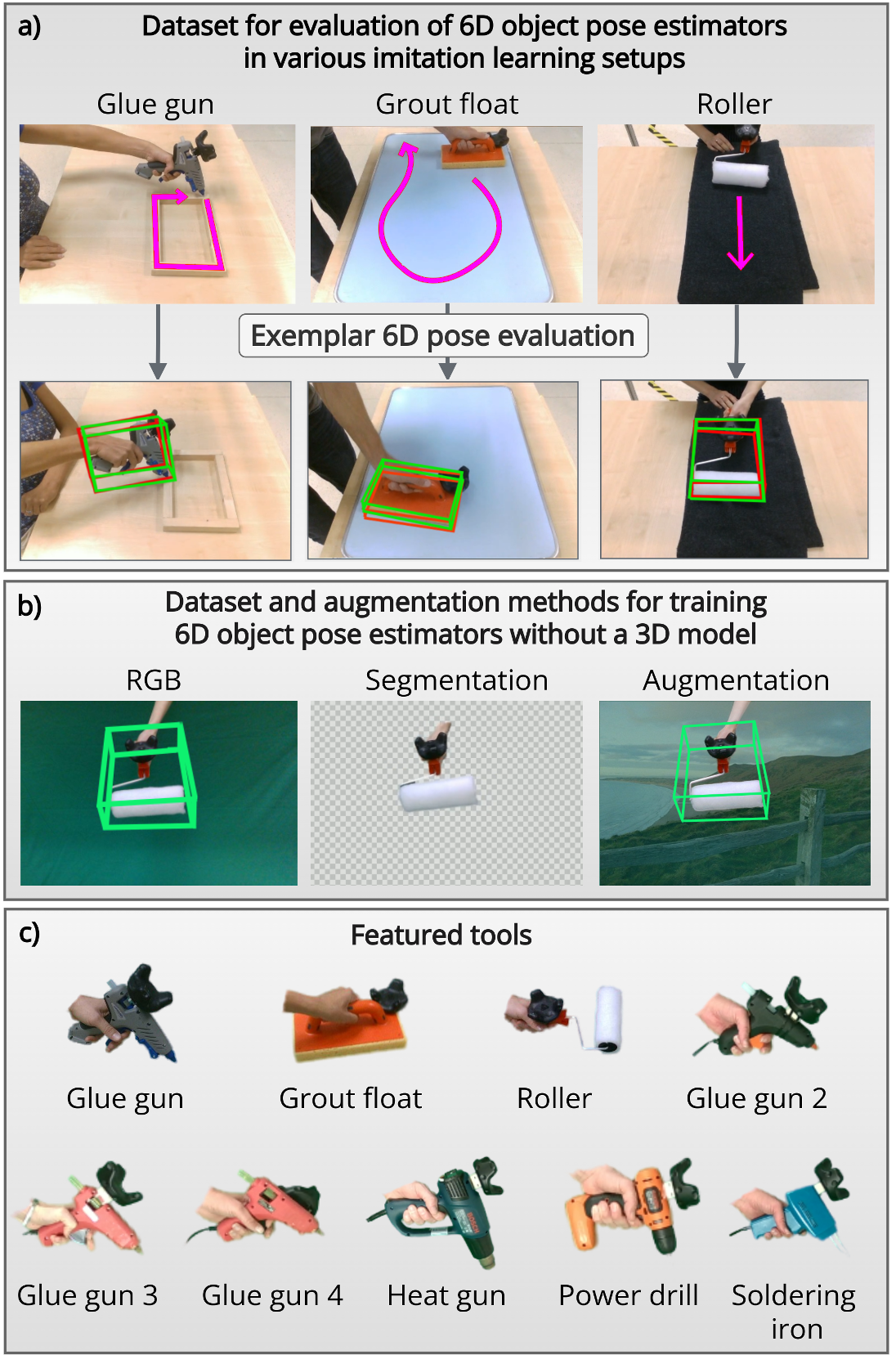}
    \caption{
\dataset dataset for \sixd object pose estimation of hand-held tools in real-world manipulation tasks in uninstrumented environments. a) \test{} dataset for benchmarking \sixd object pose estimation methods, and b) \train{} dataset for training \sixd pose estimators without a 3D model of the tool; c) the \numtoolstext tools are in both datasets. The bounding boxes visualize the predicted (red) and reference (green) object poses.}
    \label{fig:front_picture}
       \vspace{-1.5em}
\end{figure}

Acquiring data from videos of humans demonstrating a task has several potential advantages over other approaches such as kinesthetic teaching~\cite{manschitz2015learning}, teleoperation~\cite{akgun2012novel}, or motion tracking systems with markers on the objects or human body parts~\cite{ehrenmann2001sensor}. First, such a setup can provide more detailed information about the interaction of the tool with the environment. Second, it is much easier for a skilled worker to perform the task in the usual way rather than to demonstrate it by holding a robot arm. Third, such a setup enables easier transfer to different robotic platforms. Finally, vast amounts of visual data are already available (\eg instructional videos on YouTube) and can be used for learning in industrial, household, and similar settings. 

However, there are also several critical challenges that need to be addressed. First, one has to deal with the fact that hand-held objects are partially occluded by the demonstrator, exhibit various symmetries, and lack a distinctive texture. These characteristics make training of \sixd object pose estimators difficult. It is also hard to estimate a priori whether the tracking accuracy will be sufficient for the robotic task at hand. Second, to guarantee a reasonably short setup time (\eg data collection, processing, and annotation), the \sixd object pose estimator must be trainable on a limited amount of demonstration data. Finally, the 6D pose estimator should preferably work without a 3D model of the tool. However, current model-based pose estimation methods often require high-quality object models, which are difficult to acquire in real-world imitation learning applications.

The development of data-efficient and occlusion-insensitive \sixd object pose estimators requires datasets focused on manipulation with hand-held tools, as well as a methodology for evaluating the performance with regard to imitation learning tasks. Neither of these currently exists. Our paper addresses this problem and provides tools that help to improve the performance of \sixd object pose estimation in such challenging cases. Our main contributions include:
\begin{enumerate}[leftmargin=*]
\item   We have collected, annotated, and published a real-world hand-held tool manipulation dataset, called  \dataset ~\cite{imitrobdataset} (see Fig.~\ref{fig:front_picture}). 
The dataset consists of \rgbd videos with ground truth annotations of the tool \sixd poses and bounding boxes. The pose annotations are generated using the HTC Vive motion tracking system, and the bounding boxes are derived from the tracked pose and a tracing-based object surface estimation. The \dataset dataset contains \numtools hand-held tools
(four glue guns, \groutfloat, \roller, \heatgun, \powerdrill, \solder),
manipulated by 4 demonstrators, using left/right hand, and recorded from 2 camera viewpoints. The test part of the dataset (100\,332 images, \test{}), includes videos of \numtasks manipulation tasks in realistic environments. 
The training part of the dataset (83\,778 images, \train{}), contains random motion of the tools in front of green background. 
\item   We provide a methodology (accompanied by a software package~\cite{imitrobgit}) to collect ground truth training data for new objects or manipulation tasks in an affordable way. The methodology can also be used to introduce further variability into the dataset. The data acquisition methodology does not require a CAD model of the tracked tool to obtain the ground truth \sixd pose. For application in industrial environments, the methodology requires only simple modifications (\eg attaching a tracker to the tool), which we regard as essential for practical use of imitation learning in real-world set-ups, e.g.~in industrial environments.

\item   To illustrate how the \dataset dataset can be used to compare the performance of various algorithms for hand-held object pose estimation, we trained and evaluated the accuracy of a selected \sixd object pose estimator (DOPE \cite{tremblay2018deep}).

\item  We demonstrate how the generalization capabilities of the \sixd object pose estimator can be enhanced by augmentation of the \train{} dataset. For this purpose, we compared several data augmentation techniques; the best performance was achieved by a method that leverages the blending of the original and random background.
\end{enumerate}
The dataset, code~\cite{imitrobgit}, and supplementary material~\cite{imitrobsupmat} are available on the \dataset project web page~\cite{imitrobdataset}. The supplementary material contains the calibration details (Secs.~A.1-A.2), full definition of the evaluation metrics (Sec.~A.3), details about the object segmentation methods (Sec.~A.4), values of the DOPE estimator parameters (Sec.~A.5), ablation studies on the impact of image resolution, batch size, and segmentation technique (Secs.~A.6-A.7), complete results of all experiments, including metric values that did not fit into the main paper (Secs.~A.8-A.13), comparison of the model-free estimator DOPE with a model-based object pose estimator CosyPose~\cite{labbe2020cosypose} on the \powerdrill tool (Sec.~A.14), and evaluation of robustness to a change in the tracker position (Sec.~A.15).

\section{Related Work}
\label{sec:related-work}
In this section, we focus on the current datasets aimed at static \sixd object pose estimation and on those that include videos depicting manipulated objects for imitation learning. We also mention the state-of-the-art methods in \sixd object pose estimation from RGB and \rgbd images or videos.

\subsection{\sixd object pose estimation}
Motivated by applications in robotics, \sixd object pose estimation has recently attracted significant attention~\cite{tremblay2018deep,skoviera2019teaching,sahin2020review}.
In the case of richly textured objects, methods based on matching of local invariant features such as SIFT \cite{munich2006sift} or SURF \cite{bay2006surf} produce reasonable results. Unfortunately, many hand-held tools are not richly textured.
The more complicated \sixd estimation of textureless objects can be handled by models based on Convolutional Neural Networks. Methods such as ~\cite{xiang2017posecnn,wang2019densefusion} use CNNs to directly regress \sixd object pose.
In another approach, methods like \cite{li2018deepim,zakharov2019dpod,park2019pix2pose,pitteri2019cornet,hodan2020epos, he2020pvn3d, rad2017bb8} predict the correspondences between the 2D input image and either a 3D model of an object or specific keypoints on an object, which  are  then  used  to  compute  the object \sixd pose  via  the PnP algorithm. The DOPE algorithm \cite{tremblay2018deep} is a keypoint matching method that predicts  the  object's  3D  bounding box  vertices  and  centroid locations  in  the  2D  coordinate  system  of  the  input  RGB image. This approach was shown to outperform other models like the PoseCNN~\cite{xiang2017posecnn}. There are also more recent methods, such as~\cite{labbe2020cosypose,li2018deepim}, which estimate the \sixd pose based on the alignment of 3D object models with the input images. 
However, these ``render-and-compare" methods require a known 3D model of the object.
Obtaining such accurate 3D models quickly is a nontrivial task in real-world imitation learning scenarios. Hence, we choose DOPE~\cite{tremblay2018deep} as our exemplar 6D pose estimation method as it does not require a 3D model of an object to estimate the pose. Instead, only visual data and reference 6D pose data are needed. DOPE is thus more suitable for imitation learning setup as it
only requires the user to record a few short training videos with the hand-held tool.

\subsection{Datasets for \sixd object pose estimation}
One of the frequently used static datasets for object pose estimation is Linemod \cite{hinterstoisser2012model}, which consists of 15 textureless household objects with annotations and a test set that includes these objects in cluttered scenes. An extended version Linemod-Occluded \cite{brachmann2014learning} introduces a more challenging occluded testing scenario. The T-LESS dataset \cite{hodan2017t} features 30 objects from an industrial environment, which lack an easily discriminative texture and are symmetrical along one or more axes. Another industry-oriented dataset is ITODD \cite{drost2017introducing}, but its  reference annotations are not available for the test images. The recent YCB-M dataset consists of real-world static scenes recorded using 7 different cameras \cite{grenzdorffer2020ycb}. 
In all of the above-mentioned datasets, the objects are static and captured by a camera moving around the object at an approximately constant distance. However, for more realistic real-world \sixd object pose estimation, it is beneficial to train the estimators on datasets depicting manipulated objects. Creating such datasets is even more technically challenging. Hence, the existing datasets are small or employ methods that simplify the annotation task \cite{marion2018label,xiang2017posecnn}. 
For example, the authors of the YCB-Video dataset \cite{xiang2017posecnn} avoided full manual annotation by keeping the recorded objects at fixed positions and moving the camera only, leading to a high correlation of the objects' relative poses throughout the data. 
Compared to the Linemod or YCB-Video datasets, we focus on specific tasks and tools typical for industrial manufacturing environments. 
A real human activity RGB dataset focused on task-oriented grasping \cite{kokic2020learning} includes both synthetic and real RGB-D videos of manipulated objects. However, the annotations are mainly focused on the hand joint positions, and only a small proportion of the objects are provided with their meshes and 3D poses.
Probably most related to ours is the dataset \cite{krull2014}, which consists of three objects recorded with the Kinect sensors. 
Our dataset differs in the three main aspects. First, we obtain the reference \sixd pose from an HTC Vive controller attached to the object, whereas \cite{krull2014} used manual annotation. Second, \cite{krull2014} provides 3\,187 images in total,
whereas our whole dataset contains more than 184\,000 images. Third, \cite{krull2014} lacks variability across several subjects, left and right hand, different camera views, tasks, or clutter in the scene. 
Furthermore, \cite{krull2014} expects a known 3D model of the object and thus cannot be used for the training of 3D model-independent estimators.
We are unaware of any other \sixd object pose estimation video dataset besides ours that would enable the evaluation of trained models for so many different
types of generalization, \ie across different
human operators, left-handed and right-handed manipulation, task variations, camera viewpoints, occlusions, and backgrounds.
To demonstrate its utility, we measure the impact of each of these challenges on the accuracy of \sixd object pose estimation provided by the DOPE algorithm \cite{tremblay2018deep}.

\begin{figure}
  \begin{center}
      \subfloat[Data acquisition setup\label{fig:experiemental_setup}]{
      \includegraphics[width=0.65\linewidth]{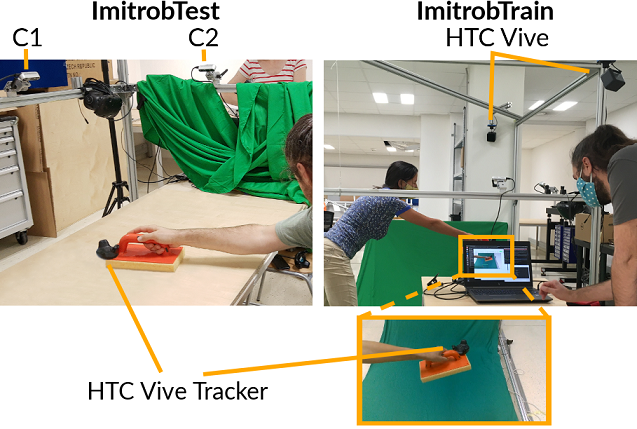} }
    \begin{minipage}{0.3\linewidth}
    \vspace{-9em}
    \subfloat[Tracing of a tool\label{subfig:trace-tool}]{
      \includegraphics[trim={0cm 0.1cm 0cm 0.1cm},clip,width=1\linewidth]{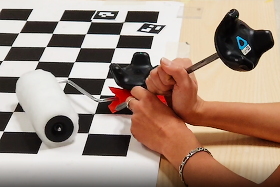} 
    }\\
    \subfloat[Tool bounding box\label{subfig:trace-bb}]{
      \includegraphics[trim={8cm 6cm 10cm 7cm},clip, width=1\linewidth]{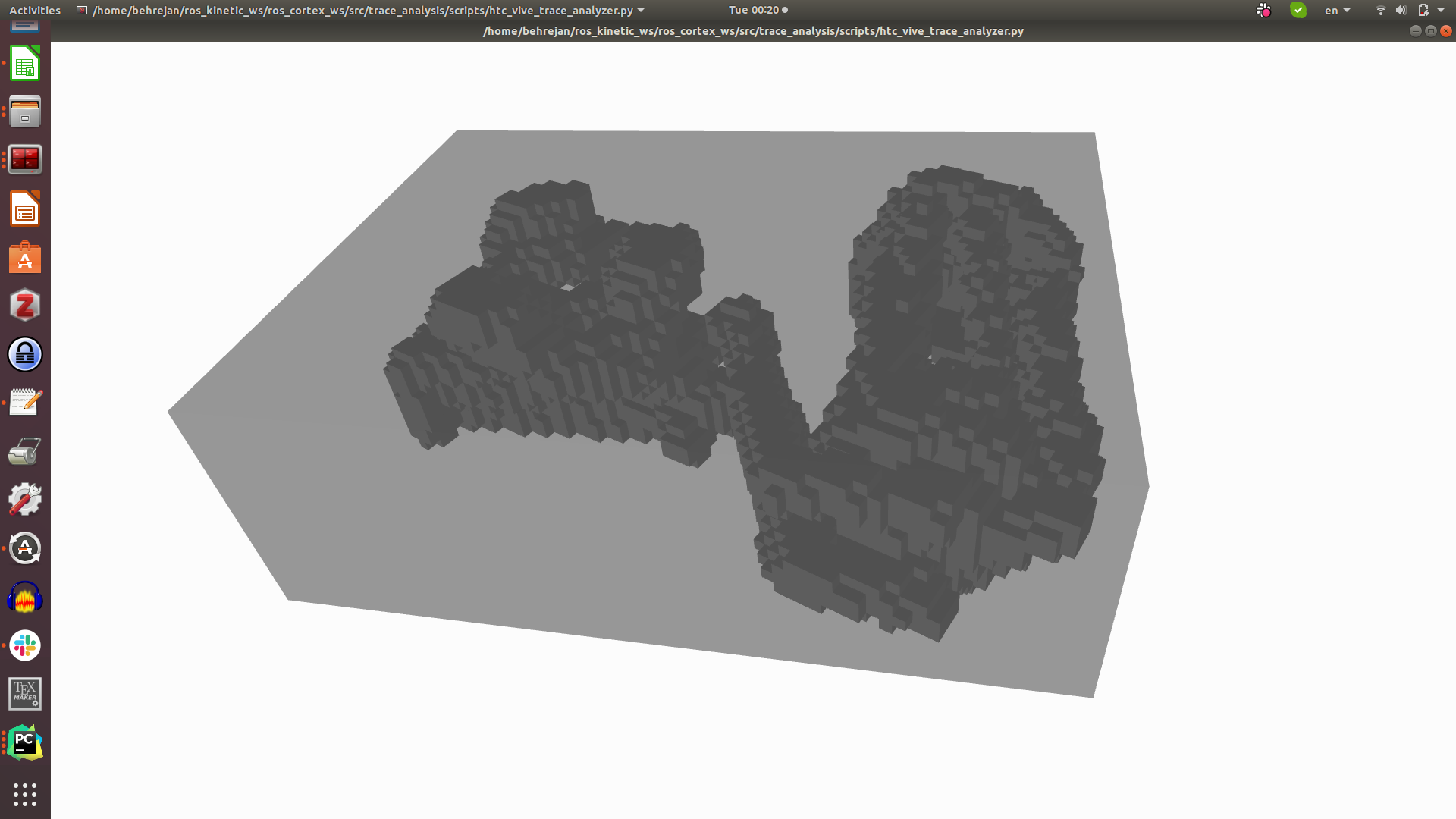}
    }
      \end{minipage}
  \end{center}
    \caption{
The experimental setup for collection of \test{} and \train{} datasets. a) The setup consists of \rgbd cameras, HTC Vive lighthouses, and a tracker attached to the tool. 
    b) The surface calibration process. c) The resulting voxel grid (dark gray) and bounding box (light gray).
    }    \label{fig:exp_setup}
       \vspace{-1.5em}
\end{figure}

\begin{figure*}[t]
    \centering
    \includegraphics[width = 0.99\linewidth]{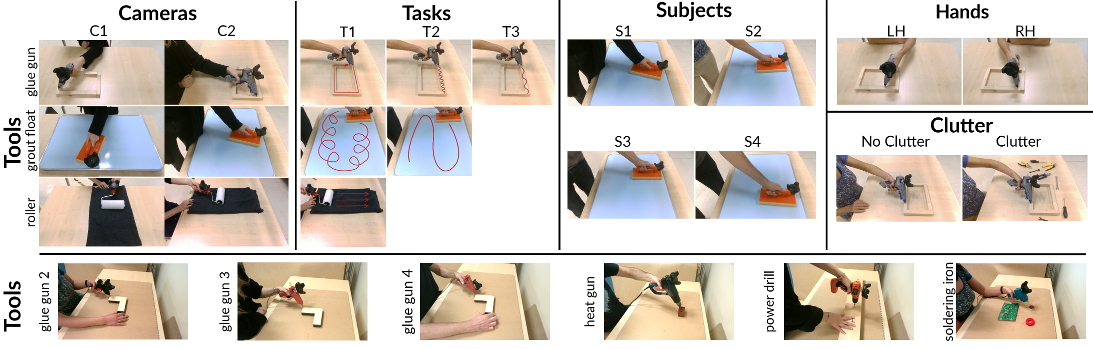}
    \caption{
Overview of the variability of setups provided in our \test{} dataset.
Tools (rows): \gluegun, \groutfloat, \roller, \gluegunII, \gluegunIII, \gluegunIIIdtp, \heatgun, \powerdrill, \solder.
Cameras: front (C1) and right hand side (C2), synchronized in the \test{} dataset. Tasks: frame (T1), dense wave (T2), and sparse wave (T3) for \gluegun; round (T1) and sweep (T2) for \groutfloat; press (T1) for \roller.
Subjects: four demonstrators (S1, S2, S3, S4).
Hands: left (LH) and right (RH) hand.
Clutter: workspace with only the gluing frame (No Clutter, default) and with other objects on the table (Clutter).}
    \label{fig:dataset_testdata}
       \vspace{-1em}
\end{figure*}

\section{Data acquisition setup}
\label{sec:experimental_setup}

The basic data acquisition setup (see Fig.~\ref{fig:experiemental_setup}) consists of a desk with two Intel RealSense D455 \rgbd cameras and an HTC Vive VR set. The data from all sensors are broadcast as Robot Operating System (ROS~\cite{quigley2009ros}) messages and stored in ROS bag files. The cameras produce 848$\times$480 \rgbd images at 60 Hz, and the HTC Vive produces \sixd poses at 30 Hz. For the data collection, each task was performed on a table with task-related or clutter objects (see Fig.~\ref{fig:dataset_testdata}). 

\subsection{Sensor setup calibration and data synchronization}
\label{sec:setupCalibration}
For the cameras, we estimated the intrinsic parameters and the radial and tangential distortion coefficients from several views of the chessboard calibration pattern using the OpenCV library~\cite{opencv_library}.
The extrinsics were calibrated from a single view of the chessboard pattern. The origin of the chessboard (world) coordinate system $O_\mathrm{w}$ was defined in one of the chessboard corners, and the camera poses relative to $O_\mathrm{w}$ were estimated by solving the PnP problem in combination with the RANSAC algorithm. To calibrate the HTC Vive coordinate frame $O_\mathrm{htc}$ (in one of the lighthouses marked as HTC Vive in Fig.~\ref{fig:experiemental_setup}) to the chessboard coordinate frame $O_\mathrm{w}$, spherical motion patterns centered at different chessboard corners $p_\mathrm{w}$ were recorded using a tracked pointing device (tracker mounted on a pointed metal rod, shown in Fig.~\ref{subfig:trace-tool}). More technical details of the calibration are presented in~\cite{imitrobsupmat}. The average deviation (residual $r_{\operatorname{avg}}$) of the acquired center points from the regular chessboard grid pattern (acquired from the cameras) was below 2\,mm for all experiments. The HTC Vive pose data were interpolated to calculate the reference object poses for the times when the individual camera images were captured.  
When the time difference between consecutive HTC Vive frames is longer than 100\,ms, the corresponding camera images are discarded to ensure sufficiently accurate ground truth data.

\subsection{HTC Vive tracker to tool calibration}
\label{sec:calibTool}
To be able to provide the reference bounding boxes for the \dataset dataset, we first have to find the bounding boxes of the manipulated objects relative to the tracker.  To find the object dimensions relative to the tracker, we traced the tool and tracker surfaces with a tracked pointing device while recording the positions of both trackers (see~Fig.~\ref{subfig:trace-tool}). Contour tracing for surface reconstruction was described in \cite{Hoppe_DeRose_Duchamp_McDonald_Stuetzle_1992}. The acquired surface points are filtered (points that are likely not part of the surface are removed), and a voxel grid with the dimensions of the object is created. Finally, we calculate a minimal bounding box aligned with the tracker's $z$-axis using the trimesh library \cite{trimesh} while evaluating volumes for different rotations. Fig.~\ref{subfig:trace-bb} visualizes the resulting voxel grid and bounding box for the \roller. Note that a small systematic error in the computed bounding boxes should not affect the performance of \sixd pose estimators because the training and testing are executed using the same bounding box calibration.
The accuracy of the pose annotations is mainly determined by the HTC Vive dynamic accuracy, which was evaluated in~\cite{Borges_Symington_Coltin_Smith_Ventura_2018} as typically around 1\,mm. The details of the whole procedure and its accuracy are described in~\cite{imitrobgit} and~\cite{imitrobsupmat}. 

\section{The imitrob dataset}
\label{sec:dataset}

Our \dataset dataset provides annotated videos of manipulation tasks with hand-held tools in settings simulating a controlled factory environment.
The motivation for the tools used in the \dataset dataset
comes from actual industrial cases.
For instance, glue guns are used in the production of aerospace equipment for airplanes, including baggage bins, trolleys etc., among many other applications.
Due to the large variability in this equipment, a large amount of repetitive manual labor is involved, which is very difficult to automate.
Another example is the sealing of plastic foil in car doors, where rollers are used to press the foil against the metal frame on which glue has been applied.
In contrast, there is less need for imitation learning for tools such as saws and screwdrivers, which are typically part of specialized robot end-effectors and thus already commonly used in many robotic applications.

We see the following three main usages of the provided dataset and methods: 1) Benchmarking \sixd pose estimation methods for hand-held tools in manipulation tasks;
2) Methodology for data acquisition and \sixd pose estimator training for new tools/tasks;
and 3) Guideline for collecting more extensive datasets and benchmarking \sixd object pose estimators on tasks with hand-held tools, e.g.~in imitation learning.

The \dataset dataset consists of:
1) \test{} dataset and evaluation metrics for benchmarking 6D object pose estimation methods and 2) \train{} dataset and augmentation methods for training 6D pose estimators that do not require a 3D model of the object.
The following sections describe these components in detail. The dataset and methods can be downloaded at~\cite{imitrobdataset}.

\subsection{\test{}: Benchmarking dataset}
\label{sec:testDataset}

The \test{} dataset (see Fig.~\ref{fig:dataset_testdata}) provides real-world benchmarking data for \sixd object pose estimation in an imitation learning setup.
It enables the evaluation of various setup combinations that one typically expects in the case of imitation learning in industrial settings. These variations include the manipulated tool, performed task, camera viewpoint, demonstrating subject, hand used for manipulation, or presence of clutter in the scene.
In total, there are 208 different tool/task/camera/hand/demonstrator/clutter combinations in the \test{} dataset.
Inspired by common trajectory-dependent industrial tasks, we focus on the scenario where the robot is observing a manipulation of a tool by a human operator in order to imitate the demonstrated trajectories.
The operator holds the tool in one hand and performs various tasks, such as applying hot glue with a \gluegun along various trajectories, polishing a surface with a \groutfloat, or flattening a cloth with a \roller. 
To learn from such demonstrations, the robot has to identify the \sixd pose of the tool. 

\paragraph{Objects and environment setups} The \dataset dataset features \numtoolstext tools (\gluegun, \groutfloat, \roller, \gluegunII, \gluegunIII, \gluegunIIIdtp, \heatgun, \powerdrill, \solder), four demonstrators (subjects S1-S4), and manipulations by the left (LH) and right (RH) hand.
The \sixd poses of the tools were measured by the HTC Vive (see Sec.~\ref{sec:calibTool}), and the image data were recorded using two \rgbd cameras from the front (C1) and right-hand side (C2) viewpoints
(see Fig.~\ref{fig:experiemental_setup}).
While we do not utilize the depth component in this work, it is included in the published dataset, along with the raw data and the code for custom data extraction.
We also included challenging, textureless and small tools. The dataset contains multiple \gluegun tools to enable testing how pose estimators generalize to different objects of the same type; while \gluegun, \gluegunII, and \gluegunIII differ in color and size, \gluegunIII and \gluegunIIIdtp differ in the position of the HTC Vive tracker (on top vs.~left side, respectively).
The \powerdrill provides a 3D model in the YCB Object set~\cite{calli15ycb}, which allows evaluation of model-based object pose estimation methods on this tool.

\begin{figure}
  \begin{center}
    \includegraphics[width=0.99\linewidth]{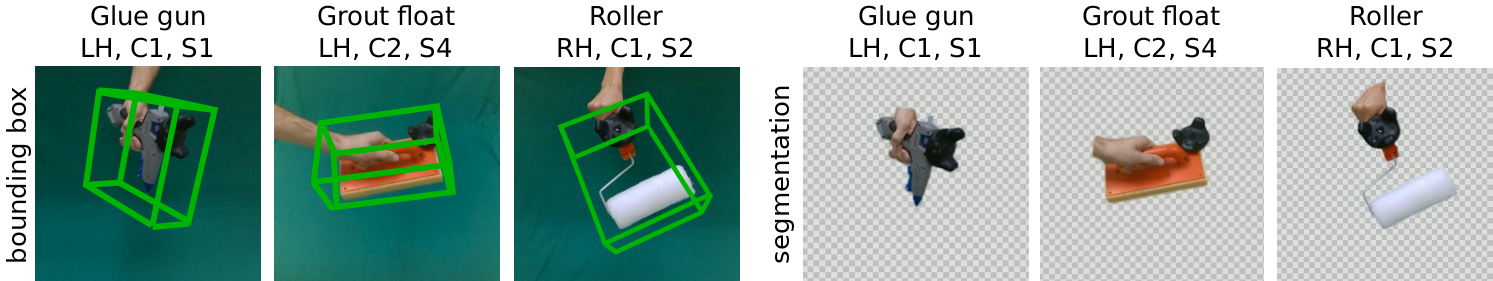}  
  \end{center}
    \caption{Example frames from our \train{} dataset.
    \textit{Left}: Bounding box of the tool computed from the HTC Vive data.
    \textit{Right}: Segmentation of the tool and hand computed by the \maskfba method.
    ``LH, C1, S1'', for example, denotes left hand, front camera, and first subject (see Sec.~\protect\ref{sec:trainDataset}).
    The images are cropped to show finer details.
   }      \label{fig:dataset_traindata}
   \vspace{-1em}
\end{figure}

\paragraph{Tasks} The \test{} dataset contains \numtaskstext tasks with different tool trajectories: three for \gluegun, two for \groutfloat, and one for each other tool (see Fig.~\ref{fig:dataset_testdata}~Tasks).
In addition, the \gluegun frame task was recorded in two environments: with only the gluing frame on the table (\noclutter, default) and with a clutter of other objects around the gluing frame (\clutter) (see Fig.~\ref{fig:dataset_testdata}~Clutter).
Each task was performed by all four demonstrators (S1-S4) to simulate the variability of tool manipulation by humans.
The dataset can thus be used for learning
task-specific motions from human demonstration.
\paragraph{Labeling of the data} All RGB-D images collected in the \dataset dataset are accompanied by a reference \sixd pose of the tool.
The \sixd poses were acquired from HTC Vive at 30Hz frequency and interpolated to match the timestamps of the camera frames (see Sec.~\ref{sec:setupCalibration} for the HTC Vive calibration details).
The \test{} dataset contains 100\,332 annotated frames.

\paragraph{Evaluation metrics}
To evaluate the performance of \sixd object pose estimators on the \test{} dataset, we use the following three metrics (further details are available in Sec.~A.3 of the supplementary material~\cite{imitrobsupmat}):
\begin{enumerate}[leftmargin=*]

\item The $\add$ pass rate ($\add_t$) measures the percentage of predictions ($P$) with $\add$ value lower than a selected threshold ($t$):
\begin{equation}\label{eq:addt}
    \add_t = \frac{|\{P | \add \leq t\}|}{|\{P\}|}  \cdot 100\% \ ,
\end{equation}
where $\add$ \cite{tremblay2018deep} is the average distance between the corresponding predicted ($p_{\oppre}^{i}$) and reference ($p_{\opref}^{i}$) vertices ($p^1,\dots,p^8$) and centroid ($p^9$) of the object bounding box:
\begin{equation}\label{eq:add}
    \add = \frac{1}{9}\sum_{i=1}^{9} ||p_{\oppre}^{i} - p_{\opref}^{i}||_2 \ .
\end{equation}
A higher $\add_t$ value for a given threshold $t$ indicates a better prediction accuracy of the object 3D bounding box.
The $\add_t$ metric is useful in imitation learning where we are interested in the absolute error regardless of the size of the manipulated object.
For comparison of models trained with ($\add_t^{\ratioaug}$) and without ($\add_t^{\rationoaug}$) augmentation, we use the ratio of their respective $\add$ pass rates:
\begin{equation}\label{eq:add_ratio}
    \add_t^{\ratio} = \frac{\add_t^{\ratioaug}}{\add_t^{\rationoaug}}\ .
\end{equation}
A higher $\add_t^{\ratio}$ value indicates a bigger benefit of the augmentation.

\item The rotation error ($E_{\rot}$) measures the angle between the predicted ($R_{\oppre}$) and reference ($R_{\opref}$) object orientations:
\begin{equation}\label{eq:rot_error}
    E_{\rot} = \arccos{\left(\frac{\operatorname{trace}({R_{\oppre}}^{-1} R_{\opref}) - 1}{2}\right)} \ .
\end{equation}
A lower $E_{\rot}$ value corresponds to a better estimate of the object orientation.

\item The translation error ($E_{\tra}$) measures the distance between the predicted ($t_{\oppre}$) and reference ($t_{\opref}$) object positions:
\begin{equation}\label{eq:tra_error}
    E_{\tra} = ||{t_{\oppre} - t_{\opref}}||_2 \ .
\end{equation}
A lower $E_{\tra}$ value indicates a better localization of the object in space.
%
\end{enumerate}

\begin{figure}[t]
    \begin{center}
\includegraphics[width=1\linewidth]{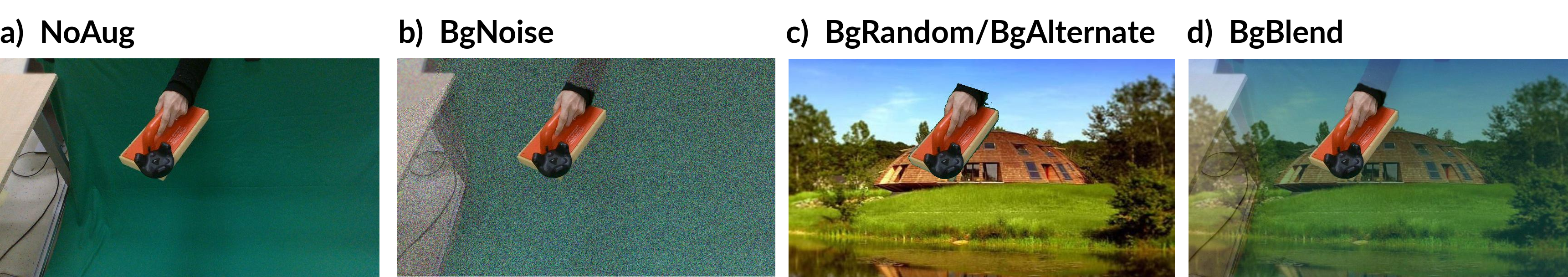}
  \end{center}
    \caption{
    Augmentation of a frame from the \train{} dataset.
    a) Original image (\noaugmentation)
    and background augmentation by
    b) \bgnoise, c) \bgrandom or \bgalternate,
    and d) \bgblend.
}   \label{fig:augmentations}
\vspace{-1em}
\end{figure}

\subsection{\train{}: Training dataset}
\label{sec:trainDataset}

The \train{} dataset (see Fig.~\ref{fig:dataset_traindata}) is designed for training \sixd object pose estimation methods that do not require a 3D model of the object.
Instead of creating a complex 3D model, the dataset captures the tools in various orientations to provide sufficient viewpoint variability for \sixd object pose training.
Each tool was moved randomly in one hand for a short time (20-40\,s) to simulate the range of possible \sixd poses during tasks. We designed this process to allow for a time and cost efficient (\eg without the requirement for manual annotation) extension of the dataset to new tools, as \sixd pose estimation methods typically require object-specific training sets.

Similarly to the \test{} dataset, the \train{} dataset contains the same \numtoolstext tools, four demonstrators (S1-S4), left and right hand (LH and RH), and two cameras (C1 and C2).
In total, there are 144 different tool/camera/hand/demonstrator combinations in the \train{} dataset.
In contrast to the \test{} dataset, which contains task-specific motions and environments, in the \train{} dataset the tools were randomly rotated in front of a green background, which enables automatic object segmentation for background augmentation (see~\cite{imitrobdataset}).
We include the segmented data as well as the reference \sixd object pose from HTC Vive in the published \train{} dataset.
The \train{} dataset contains 83\,778 frames.
The size of the dataset corresponds to the application area of imitation learning in industrial settings, where real data has to be collected from human demonstrators.

\paragraph{Augmentation methods} We provide a collection of data augmentation methods suitable for the \train{} dataset (code at~\cite{imitrobgit}).
The augmentation significantly increases the size of the training set and robustness of the trained model to the variability of the test environment.
First, we leverage the green background to segment the tool and hand by thresholding (\maskthresholding), followed by $F$, $B$, Alpha Matting \cite{fbamatting2020} (\maskfba) (see Fig.~\ref{fig:dataset_traindata}, for details see Sec.~A.4 in \cite{imitrobsupmat}).
Then we apply a random crop (constrained to keep all vertices of the 3D bounding box inside the image) and horizontal flip, and one of the following background randomization techniques.
\bgrandom replaces the background with a random image \cite{li2022motionforcesfromvideo}. 
The other three methods keep the original background for 25\% of the training images, and for the remaining 75\% \bgalternate replaces the background with a random image, \bgblend blends the background with a random image,
and \bgnoise blends the background with random color noise.
In our experiments, the random images were sampled from the miniImageNet~\cite{vinyals2016matching} dataset of 60\,000 images.
Fig.~\ref{fig:augmentations} shows a training image without augmentation (\noaugmentation) and after augmentation by these methods.

\section{Exemplar \sixd object pose estimation performance}
\label{sec:experimentsResults}

In the following experiments we demonstrate the utility of the \dataset dataset for \sixd object pose estimation training and testing.
For this purpose we use the \sixd object pose estimator DOPE \cite{tremblay2018deep} (see Sec.~A.5 in~\cite{imitrobsupmat} for implementation details).
We train all models on the \train{} dataset (see Sec.~\ref{sec:trainDataset}) and evaluate their performance on the \test{} dataset (see Sec.~\ref{sec:testDataset}).
We compute the ADD pass rates for threshold $t = $ \fivecm ($\add_5$),
as well as the rotation ($E_{\rot}$) and translation ($E_{\tra}$) errors.
First, we present an ablation study to motivate our choice of the data augmentation method (computed for \gluegun, \groutfloat, and \roller).
Then we focus on the ability of the \sixd object pose estimator to generalize to various training/test setups, including combinations of front/side camera, left/right hand, subjects (all computed for \gluegun, \groutfloat, and \roller), and background clutter (\gluegun task frame).
Finally, we report performance for each tool and manipulation task.
Full results, including $\add$ values
for thresholds $t = $ \twocm ($\add_2$) and \tencm ($\add_{10}$),
ablation studies on the impact of image resolution, batch size, and segmentation method, comparison of model-free DOPE and model-based CosyPose object pose estimation methods on the \powerdrill tool,
as well as robustness to a different tracker position between tools \gluegunIII and \gluegunIIIdtp are available in the supplementary material~\cite{imitrobsupmat} (see Secs.~A.6-A.15).

\begin{table}
\caption{ Comparison of
data augmentation methods (see Sec.~\protect\ref{sec:trainDataset}).}
\setlength{\tabcolsep}{2pt}
\centering
\begin{tabular}{cccccc}
$\add_5$ & \multicolumn{1}{c}{ \noaugmentation } & \multicolumn{1}{c}{ \bgnoise } & \multicolumn{1}{c}{ \bgrandom } & \multicolumn{1}{c}{ \bgalternate } & \multicolumn{1}{c}{ \bgblend } \\
\hline
\gluegun  & 17.0  & 18.9 &  36.5  & 45.4  & {\bf 57.8}  \\
\groutfloat  & 45.4  & 43.6  & 60.4  & 71.8  & {\bf 73.9}  \\
\roller  & 25.3  & 26.3  & 39.8  & {\bf 52.2}  & 48.6  \\
\hline
average  & 29.2  & 29.6  & 45.6 & 56.5  & {\bf 60.1} \\
\hline
\hline
$\add_5^{\ratio}$  & - & 1.0 & 1.6 & 1.9 & {\bf 2.1} \\
\end{tabular}
\label{tab:aug_methods}
\vspace{-0.5em}
\end{table}

\subsection{Ablation experiments}

\paragraph{Benefits of data augmentation} Table~\ref{tab:aug_methods} shows the effect of the background augmentation methods from Sec.~\ref{sec:trainDataset} on the performance of the \sixd object pose estimator DOPE.
For the object segmentation step, we use the \maskfba method, which outperforms the simple \maskthresholding (see Sec.~A.7 in \cite{imitrobsupmat}).
Real-world images (\bgrandom, \bgalternate, \bgblend) clearly outperform color noise (\bgnoise) as a random background for augmentation.
Moreover, it is beneficial to keep the original background for a portion (in our case 25\%) of training images (\bgalternate, \bgblend) rather than to replace the background everywhere (\bgrandom).
The best results were achieved
by \bgblend, which (after the random crop, horizontal flip, and segmentation by the \maskfba method) blends the original background with a random image for 75\% of the training images and keeps the original background for the remaining 25\% of the training images.
Compared to training without augmentation (\noaugmentation), the use of \bgblend augmentation increased the $\add_{5}$ accuracy more than twofold (from $29.2\%$ to $60.1\%$).
Thus, in all other experiments, we use the \bgblend background augmentation.

\paragraph{Generalization across camera viewpoints}
\label{sec:acrossCamera}
In this experiment, we study the robustness of the \sixd object pose estimator with respect to the camera viewpoint.
The \dataset dataset contains one front camera (C1) and one right-hand side camera (C2). 
Table~\ref{tab:generalizationResults} compares results for the following scenarios:
a) \samecamera: training and testing on the same camera;
b) \othercamera: training on one camera and testing on the other;
c) \bothcameras: training on both cameras.
The accuracy of using a different camera viewpoint between training and testing was
very low;
on average, the results were better for the transfer from C1 to C2 ($\add_{5} = 1.1\%$) than for the transfer from C2 to C1 ($0.1\%$). The accuracy was significantly higher when the camera used for testing was included in the training. 
The best results for evaluation on C1 were achieved by models trained on both C1 and C2 (\bothcameras), while the best results for evaluation on C2 were achieved by models trained only on C2 (\samecamera).

\begin{table}
\caption{Generalization across camera viewpoints, left/right hand, and demonstrators (see Sec.~\protect\ref{sec:experimentsResults})
for tools \gluegun, \groutfloat, and \roller (average values).}
\label{tab:generalizationResults}
\setlength{\tabcolsep}{2pt}
\centering
\begin{tabular}{cccc||ccc||cc}
\multirow{2}{*}{$\add_5$} & Training & \multicolumn{2}{c||}{ Test } & Training & \multicolumn{2}{c||}{ Test }& Subject & \multirow{2}{*}{ Test } \\
 & camera & C1  & C2 & hand & LH & RH & scenario & \\
\hline
\multirow{3}{*}{ \bgblend } & \samecamera  & 56.3  & {\bf 52.2}  & \samehand & 57.2 & 40.2 &\allsubjects & {\bf 58.8}\\
& \othercamera  & 0.1  & 1.1  &\oppositehand & 21.0 & 27.0  &\differentsubject & 52.0 \\
& \bothcameras  & {\bf 69.3} & 49.0  & \bothhands & {\bf 60.5} & {\bf 57.0}& \samesubject & 31.1\\
\hline
\multirow{3}{*}{ \noaugmentation } & \samecamera  & 28.7 & 41.8 & \samehand & 29.4 & 22.5 & \allsubjects & 28.9 \\
& \othercamera & 0.0 & 0.8 & \oppositehand & 13.7 & 18.4 & \differentsubject & 22.3 \\
& \bothcameras & 23.5 & 34.8 & \bothhands & 31.3 & 25.6 & \samesubject & 18.7 \\
\hline
\hline
\multirow{3}{*}{ $\add_5^{\ratio}$} & \samecamera  & 2.0  & 1.2  & \samehand & {\bf 1.9} & 1.8 &\allsubjects & 2.0 \\
& \othercamera  & -  & 1.4  &\oppositehand & 1.5 & 1.5  &\differentsubject & {\bf 2.3} \\
& \bothcameras  & {\bf 2.9} & {\bf 1.4}  & \bothhands & 1.9 & {\bf 2.2} & \samesubject & 1.7 \\
\end{tabular} 
\vspace{-1em}
\end{table}

\paragraph{Generalization across left/right hand}
\label{sec:hand_gen}
We explore the generalization of the \sixd pose estimator to manipulation of the tool by the left (LH) or right (RH) hand. 
Table~\ref{tab:generalizationResults} compares the results for 
the following cases:
a) \samehand: training and testing on the same hand;
b) \oppositehand: training on one hand and testing on the other;
c) \bothhands: training on both hands.
While training and testing on the same hand (\samehand, $\add_{5} = 48.7\%$) is clearly better than on the opposite hand (\oppositehand, $24.0\%$), using both LH and RH for training further improved the accuracy (Both, $58.8\%$).
The data augmentation was more beneficial for training on both hands (Both, $\add_5^{\ratio} = 2.1\times$) than for training only on the same (Same, $1.9\times$) or opposite hand (Opposite, $1.5\times$).
The $\add_{5}$ and $\add_5^{\ratio}$ values in this paragraph are averages across LH and RH in the test set.

\paragraph{Generalization across demonstrators}
\label{sec:subjects}
To be transferable, the \sixd pose estimation algorithm should be invariant to the subject that manipulates the tool.
We examine the generalization of the DOPE estimator across 4 different subjects (S1-S4) using
the following setups:
a) \allsubjects: train one model on all 4 subjects
(\ie train and test on S1-S4);
b) \differentsubject: train one model on 3 subjects and test it on the remaining one 
(\eg train on S1-S3 and test on S4);
and c) \samesubject: train one model for each subject
and test it on the same subject 
(\eg train and test on S1).
Table~\ref{tab:generalizationResults} averages the model accuracy for each setup across all test subjects (\ie S1-S4).
The \allsubjects setup ($\add_{5} = 58.8\%$) outperformed the \differentsubject setup ($52.0\%$), which in turn clearly outperformed the \samesubject setup ($31.1\%$).
Additionally, the augmentation improves the accuracy more for \differentsubject ($\add_5^{\ratio} = 2.3\times$) than for the \allsubjects ($2.0\times$) and \samesubject ($1.7\times$) setups. 

\begin{table}
\caption{Robustness to clutter (see Sec.~\protect\ref{sec:experimentsResults}) for tool \gluegun and task frame.}
\centering
\begin{tabular}{ccccccc}
$\add_5$ & \noclutter & \clutter \\
\hline
\bgblend & {\bf 61.8} & {\bf 61.5} \\
\noaugmentation & 22.8 & 4.9 \\
\hline
\hline
$\add_5^{\ratio}$ & 2.7 & 12.6 \\
\end{tabular}
\label{tab:clutter}
\end{table}

\begin{table}
\caption{Comparison of performance on different tools and manipulation tasks. 5~cm ADD pass rate ($\add_5$) accuracy and average rotation ($E_{\rot}$) and translation ($E_{\tra}$) errors for different tools and tasks.
Invalid detections
were excluded from the computation of $E_{\rot}$ and $E_{\tra}$.}
\label{tab:tasks}
\setlength{\tabcolsep}{8pt}
\centering
\begin{tabular}{ccccc}
\multirow{2}{*}{Tool} & \multirow{2}{*}{Task} & $\add_5$ & $E_{\rot}$ & $E_{\tra}$ \\
{ } & { } & (\%)  & (deg) & (cm)\\
\hline
\multirow{4}{*}{\gluegun} & frame &  53.3 &  11.8 & 5.0\\
& densewave &  61.9 & 5.0 & 3.6\\
& sparsewave & 66.0 &  5.0 & 3.4\\
\cline{2-5}
& average & 60.4 & 7.3 & 4.0 \\
\hline
\multirow{3}{*}{\groutfloat} & round & 74.4 &  {\bf 3.9} & 2.7\\
& sweep &  {\bf 82.7} & 4.3 & {\bf 2.2}\\
\cline{2-5}
& average & 78.6 & 4.1 & 2.5 \\
\hline
\roller & press &  50.5 &  8.7 & 3.7\\
\gluegunII & lshape & 9.0 & 38.5 & 9.9 \\
\gluegunIII & lshape & 4.7 & 40.3 & 10.2 \\
\gluegunIIIdtp & lshape & 23.4 & 20.9 & 8.4 \\
\heatgun & heating & 13.2 & 14.3 & 7.0 \\
\powerdrill & down & 59.8 & 8.0 & 3.8 \\
\solder & soldering & 12.8 & 35.6 & 9.0 \\
\hline
average & - & 34.7 & 19.8 & 6.5 \\
\end{tabular}
\vspace{-1em}
\end{table}

\paragraph{Robustness to clutter}
\label{sec:gen_clutter}
To explore the generalization of the \sixd object pose estimator to clutter in the test data, we compare its performance for the \gluegun task frame with only the gluing frame on the table (\noclutter, default) and with a clutter of other objects around the frame (\clutter) (see Table~\ref{tab:clutter}).
While the model trained without data augmentation (\noaugmentation) was clearly worse on \clutter ($\add_5=4.9\%$) than on \noclutter ($22.8\%$), the use of data augmentation not only clearly improved the performance on both subsets but also increased the accuracy on \clutter ($61.5\%$) to the same level as on \noclutter ($61.8\%$).
The $\add_5^{\ratio}$ improvement ratio was $12.6\times$ for \clutter, compared with $2.7\times$ for \noclutter, indicating a big benefit of training with data augmentation for \sixd object pose estimation in cluttered environment.

\subsection{Final results}

\paragraph{Performance on different tools and tasks}
\label{sec:tasks}

Table~\ref{tab:tasks} shows \fivecm ADD pass rates and rotation and translation errors for individual tools and tasks (see Fig.~\ref{fig:dataset_testdata}).
The tested object pose estimator performed comparably on different tasks of the same tool. The best results were achieved for \groutfloat ($\add_5 = 78.6\%$, $E_{\rot} = 4.1^\circ$, and $E_{\tra} = 2.5$\,cm), while the most challenging tools included \gluegunII and \gluegunIII (small size and large occlusions)
and \solder (textureless glossy surface). 
Overall, the average \fivecm ADD pass rate was $\add_5 = 34.7\%$ and the average rotation and translation errors were $E_{\rot} = 19.8^\circ$ and $E_{\tra} = 6.5$\,cm, respectively.

\begin{figure}
  \begin{center}
    \includegraphics[width = 0.16\textwidth]{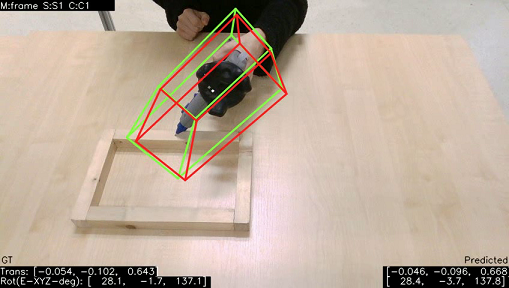}\includegraphics[width = 0.16\textwidth]{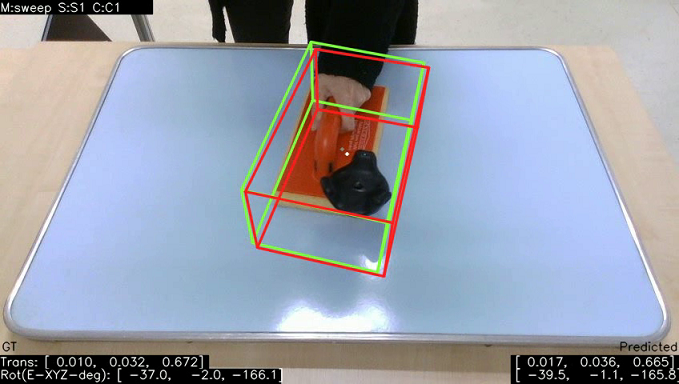}\includegraphics[width = 0.16\textwidth]{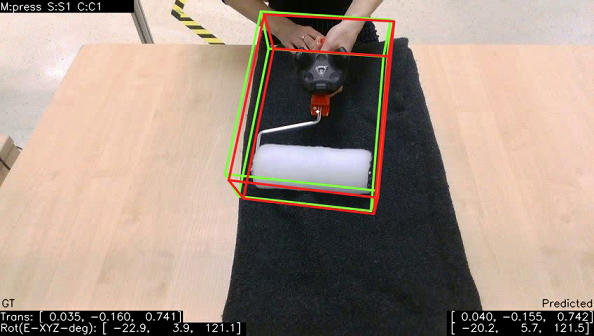}\hfill\includegraphics[width = 0.16\textwidth]{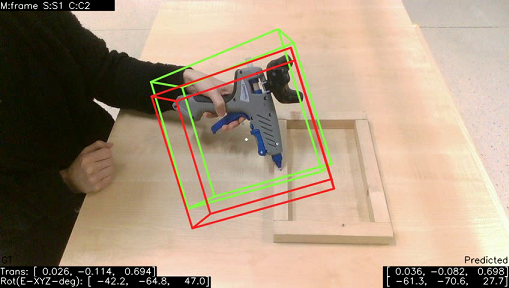}\includegraphics[width = 0.16\textwidth]{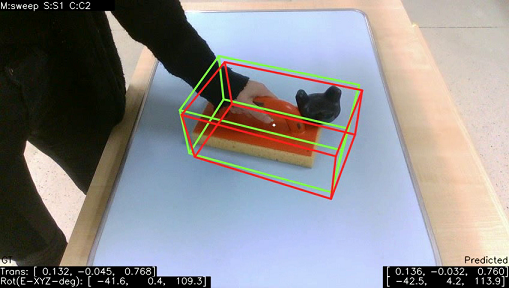}\includegraphics[width = 0.16\textwidth]{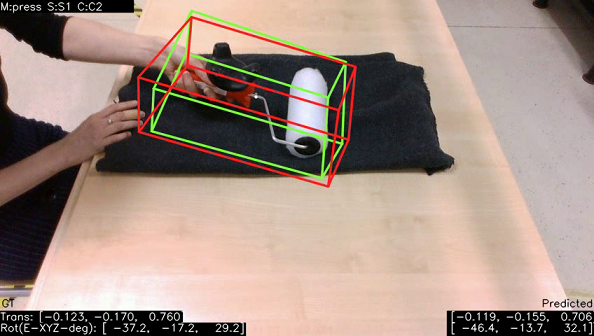}\hfill
    \end{center}
    \caption{Example qualitative results on the \test{} dataset. Comparison of the reference bounding box (green) and the bounding box predicted by the \sixd object pose estimator DOPE (red) trained on the \train{} dataset. More results are available in the supplementary video on the \dataset project web page~\protect\cite{imitrobdataset}.}
    \label{fig:quality_6dof_train_test}
       \vspace{-1em}
\end{figure}

\paragraph{Qualitative results}
Fig.~\ref{fig:quality_6dof_train_test} presents example qualitative results of the \sixd object pose estimator DOPE
trained on the \train{} dataset using the best data augmentation (\ie random crop and horizontal flip, segmentation by the \maskfba method, and background randomization by the \bgblend method, see Sec.~\ref{sec:trainDataset}) and tested on the \test{} set.
The predicted bounding box is shown in red while the reference bounding box, acquired through the camera-to-tracker and tracker-to-tool calibration (see Sec.~\ref{sec:experimental_setup}), is green.

\section{Conclusions and lessons learned}
\label{sec:conclusion}
In this paper, we address the problem of \sixd pose estimation of hand tools manipulated by human demonstrators in an industrial environment from RGB image data.
To investigate this problem, we have collected a challenging real-world benchmark video dataset (\dataset dataset) of \numtaskstext manipulation tasks with \numtoolstext different tools performed by four human demonstrators using left/right hand and recorded from two camera viewpoints (front and side).

We performed a broad range of experiments with various parts of the \train{} and \test{} datasets using the object pose estimation method DOPE~\cite{tremblay2018deep} to show the suitability of the \dataset dataset for benchmarking \sixd object pose estimation methods as well as to point out the limitations of current methods in this setup. The experiments imply that it is crucial to include in the training data the camera viewpoint used during the inference.
Manipulation by the hand opposite to the side camera led to lower occlusion of the tool and higher accuracy of the pose estimator.
The performance of models that used the same subject/camera/hand combination in both training and test data was often boosted by adding other demonstrators, camera viewpoint, or the other hand into the training set (see Table~\ref{tab:generalizationResults}).

To enhance the training data, we proposed several data augmentation methods that we provide together with the dataset. The best results were achieved by the background blending method \bgblend (see Table~\ref{tab:aug_methods}), which increased the generalization capability of the trained models in all setups.
To our knowledge, this is the first application of the background blending augmentation, previously used in image classification \cite{zhang2018mixup,yun2019cutmix}, in the \sixd object pose estimation domain.

The pose estimation accuracy correlated with the size and texture of the tool as well as with the performed task or clutter in the test environment (see Table~\ref{tab:tasks}).
The best results ($\add_5 = 78.6\%$, $E_{\rot} = 4.1^\circ$, $E_{\tra} = 2.5$\,cm) were achieved for \groutfloat, which is large and moved along a plane, while the worst accuracy was observed for small tools with textureless surface and less restricted movement.
Although the achieved accuracy of the evaluated method (DOPE) may not be sufficient for some industrial applications
, the results are promising and show that the \dataset dataset can be used to benchmark and select \sixd object pose estimation methods for various tasks based on the required accuracy. We hope that the presented dataset will trigger further development of \sixd object pose estimation methods so that learning by demonstration using only visual information will soon become a reality.

%% file: secs/supmat.tex
\title{Imitrob: Imitation Learning Dataset for Training and Evaluating 6D Object Pose Estimators\\ {[}Supplementary material{]}}

\author{Jiri~Sedlar$^{1,*}$, Karla~Stepanova$^{1,*}$, Radoslav~Skoviera$^{1}$, Jan~K.~Behrens$^{1}$, Matus~Tuna$^{2}$, Gabriela~Sejnova$^{1}$, Josef~Sivic$^{1}$, and~Robert~Babuska$^{1,3}$%
\thanks{$^{1 }$Czech Institute of Informatics, Robotics and Cybernetics, Czech Technical University in Prague, Czech Republic}
\thanks{$^{2 }$Faculty of Mathematics, Physics and Informatics, Comenius University in Bratislava, Slovakia}
\thanks{$^{3 }$Cognitive Robotics, Faculty of 3mE, Delft University of Technology, The Netherlands}
\thanks{$^{* }$Both authors contributed equally. E-mail: jiri.sedlar@cvut.cz, karla.stepanova@cvut.cz}%
}

\maketitle




This supplementary material provides additional information to paper \textit{Imitrob: Imitation Learning Dataset for Training and Evaluating 6D Object Pose Estimators} (DOI \href{https://ieeexplore.ieee.org/document/10077123}{10.1109/LRA.2023.3259735}) \cite{imitrob2023}.
Sec.~A covers the setup details and additional experimental results,
Sec.~B contains links to the dataset documentation and supplementary code and describes the intended uses of the dataset,
Sec.~C provides details on the dataset licensing and hosting, including the maintainance plan,
and Sec.~D contains a standardized datasheet for the \dataset dataset. 

\section*{A. Setup Details and Additional Experimental Results}
\label{sec:appendix}

This section provides details about the \dataset dataset acqusition setup, evaluation metrics, segmentation methods, \sixd object pose estimator setup, and all experiments, including a comparison with another object pose estimator and the impact of the tracker position.

The sensor setup calibration is described in Sec.~A.1 and the method for calibration of the HTC Vive tracker to the tool is explained in Sec.~A.2.
The evaluation metrics are defined in detail in Sec.~A.3 and the object segmentation methods used for background augmentation are described in Sec.~A.4.
The \sixd object pose estimator DOPE~\cite{tremblay2018deep} settings used for the experiments are given in Sec.~A.5.

The remaining sections contain detailed experimental results and ablation studies.
Sec.~A.6 shows the impact of image resolution and batch size on the accuracy of the \sixd object pose estimator, while Sec.~A.7 compares the impact of different object segmentation methods on the benefit of the background augmentation.
Secs.~A.8-A.13 contain complete results of the experiments evaluating different data augmentation methods, generalization across camera viewpoints, left/right hand, demonstrators, robustness to clutter, and performance on different tools and tasks, respectively.
In Sec.~A.14 we compare the model-free estimator DOPE with a model-based \sixd object pose estimator CosyPose~\cite{labbe2020cosypose} on the \powerdrill tool and in Sec.~A.15 we evaluate the performance of the \sixd object pose estimator DOPE with respect to different HTC Vive tracker positions.



\subsection*{A.1 Sensor setup calibration}
\label{sec:setup-calibration-details}
\begin{figure}[t]
  \begin{center}
    \includegraphics[width=0.48\textwidth]{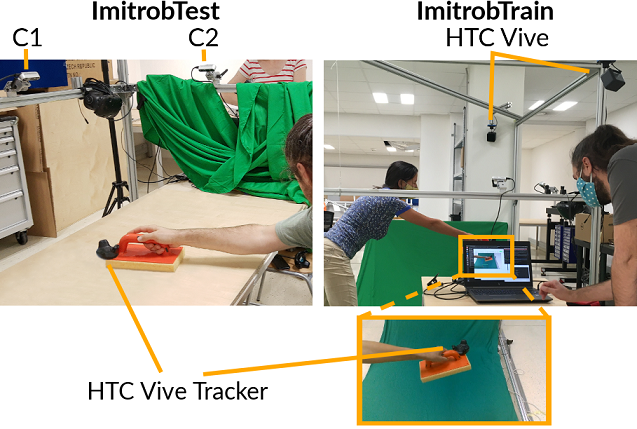}
  \end{center}
    \caption{The experimental setup for collection of \test{} and \train{} datasets. The setup consists of two \rgbd cameras (front camera C1 and right-hand side camera C2), two HTC Vive lighthouses, and an HTC Vive tracker attached to the tool.}
\label{fig:experiemental_setup_supmat}
\end{figure}

To calibrate the HTC Vive coordinate frame $O_\mathrm{htc}$ (in one of the lighthouses marked as HTC Vive in Fig.~\ref{fig:experiemental_setup_supmat}) to the chessboard coordinate frame $O_\mathrm{w}$, spherical motion patterns centered at different chessboard corners $p_\mathrm{w}$ were recorded using an HTC Vive tracker mounted on a pointed metal rod (the rod is shown in Fig.~\ref{subfig:trace-tool_supmat}).

The sphere center points $p_\mathrm{htc}$ (relative to $O_\mathrm{htc}$) were computed using orthogonal distance regression. The distances of all center points to the common plane found using RANSAC were smaller than 1\,mm, i.e.~all points lie on the flat plane of the chessboard pattern. The optimal Euclidean transformation $H$ from $p_\mathrm{htc}$ to $p_\mathrm{w}$ (i.e.~transformation between $O_\mathrm{htc}$ and $O_\mathrm{w}$) was found using the SVD algorithm \cite{sorkine2017least}. The average deviation (residual $r_{\operatorname{avg}}$) of the acquired center points from the regular chessboard grid pattern (acquired from the cameras) was below 2\,mm for all experiments. The deviation was calculated as
\begin{equation}\label{eq:calib-residual}
    r_{\operatorname{avg}}=\stackrel[i=1]{N}{\sum}\frac{\left\Vert Hp_\mathrm{htc}^{i}-p_\mathrm{w}^{i}\right\Vert _{2}+\left\Vert H^{-1}p_\mathrm{w}^{i}-p_\mathrm{htc}^{i}\right\Vert _{2}}{2N}\ ,
\end{equation}
where $N$ is the number of acquired center points and corresponding points in coordinate frames $O_\mathrm{htc}$ and $O_\mathrm{w}$ have the same index.

The final accuracy of the ground truth poses is also dependent on the ability of the HTC Vive to provide accurate and stable poses of the tracker attached to a tool with respect to $O_\mathrm{htc}$. The accuracy of HTC Vive in dynamic situations is evaluated in detail in~\cite{Borges_Symington_Coltin_Smith_Ventura_2018}.

\subsection*{A.2 HTC Vive tracker to tool calibration}
\label{sec:htc-vive-tracker-calib-details}

The method presented in this section allows finding a description of the object surface with respect to an attached motion tracker that provides reference \sixd data. 
In this paper, it is used to find the bounding boxes of the manipulated objects relative to the tracker, which in turn are used to generate the reference bounding boxes for the \dataset dataset. 

\begin{figure}[t]
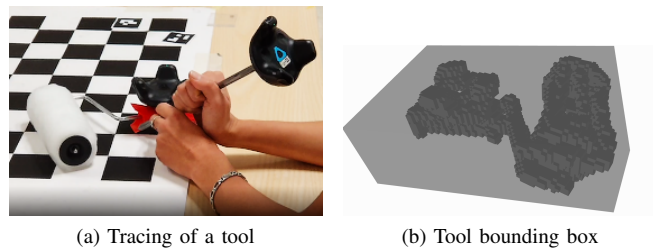

    \begin{center}
    \subfloat[Tracing of a tool\label{subfig:trace-tool_supmat}]{
      \includegraphics[width=0.47\linewidth]{imgs/tracing_roller.png} 
    }
    \hfill
    \subfloat[Tool bounding box\label{subfig:trace-bb_supmat}]{
      \includegraphics[trim={8cm 5cm 10cm 6cm},clip, width=0.47\linewidth]{imgs/roller_bb.png}
    }
    \end{center}
        \caption{Calibration of the tool with respect to the tracking device. a) The tool (\roller) surface is traced with a pointing device. b) The collected data (here 6\,364 surface trace points) is used to calculate a voxel grid for the tool (dark gray) and the final bounding box (light gray).
    }
    \label{fig:tool-calib}
\end{figure}

Note that the computed bounding boxes do not affect the performance of \sixd pose estimators because the training and testing are executed using the same bounding box calibration. The accuracy of the pose annotations is mainly determined by the HTC Vive dynamic accuracy, which was evaluated in \cite{Borges_Symington_Coltin_Smith_Ventura_2018}. Nonetheless, we look for bounding boxes that 1) contain the object, 2) align with the tracker axis, and 3) are minimal in size. In this way, the bounding boxes can be used to create the segmentation masks provided with the dataset and ensure consistent appearance in different experiments.
The tracker attachment was chosen to allow unobstructed handling of the tool and good visibility of the tracker. If possible, the tracker was aligned with the main tool axis. 
To find the object dimensions relative to the tracker, we traced the tool and tracker surfaces with a pointing device (pointed rod with another HTC Vive tracker) while recording the positions of both trackers (see~Fig.~\ref{subfig:trace-tool_supmat}). Contour tracing for surface reconstruction was described in \cite{Hoppe_DeRose_Duchamp_McDonald_Stuetzle_1992}.
We transform the $N$ recorded pointing tip points into the frame of the tool tracker to obtain a set of measurements  $P = \{p_i \subset \mathbb{R}^3\}$. The existence of a non-empty set $\hat{P} \subset P$ of outliers makes filtering of the measurements $P$ necessary. Our filtering approach based on measurement density is explained next.
For a regular grid $V$ within the axis-aligned bounding box of the traced volume, we calculate a measurement density $d_{i}$ as the number of measurements closer than a threshold $\delta$ at each grid vertex $v_i \in V$:
\begin{equation}\label{eq:density-meas}
  d_{i} = \sum_{p_i \in \Gamma} 
  \begin{cases}
                                  0 & \text{if $||p_i - v_i||_2 > \delta$ }\\
                                  1 & \text{if $||p_i - v_i||_2 \leq \delta$}\ ,
  \end{cases}
\end{equation}
where $\Gamma$ is a subset of the set of measured points $P$.
We consider the voxel centered at grid vertex $v_i$ to be part of the object surface if the density of measurements $d_i$ at the given grid cell is larger than a threshold $\zeta$, i.e.  $d_i \geq \zeta$. 
To approximate Eq.~(\ref{eq:density-meas}) for the purpose of comparing it with the threshold $\zeta$, we use a \textit{k-d Tree} to efficiently organize the measurement points $p_i \in P$ and query it for the $\zeta + 1$ nearest neighbors $P_i$ for each grid vertex $v_i$ with a cut-off distance of $\delta + \epsilon$. We evaluate Eq.~(\ref{eq:density-meas}) for $\Gamma = P_i$ to decide if $v_i$ is part of the object's surface. In short, we decide for each grid point whether it is part of the tool's surface by evaluating how many measurements are present in its vicinity. For efficiency, we check only just enough nearest neighbors to decide if the threshold was reached.

From this, we create a voxel grid with the dimensions of the object, on which we calculate a minimal bounding box using the trimesh library \cite{trimesh}.
To find instead the smallest bounding box that is aligned with the tracker's $z$-axis (second bounding box property), we rotate the occupied voxels in $0.1\degree$ steps around the $z$-axis and record the volume of each axis-aligned bounding box. The rotation with the smallest volume is then used.  
In this work, we used $\delta=0.01$\,m and a resolution of $200$ grid points per meter.
Fig.~\ref{subfig:trace-bb_supmat} visualizes the  
resulting voxel grid (dark gray) and the bounding box (light gray) for the \roller.

\subsection*{A.3 Evaluation metrics}
\label{sec:evaluation-metrics-details}

The \sixd object pose can be defined by 3D coordinates of the bounding box vertices $\mathbf{p}^1,\dots,\mathbf{p}^8 \in \mathbb{R}^3$ or by a rigid transformation $[\mathbf{R}|\mathbf{t}] \in SE(3)$, consisting of a rotation matrix $\mathbf{R} \in SO(3)$ and a translation vector $\mathbf{t} \in \mathbb{R}^3$. To evaluate the performance of a \sixd object pose estimator on the \test{} dataset, we use the following three metrics.

\paragraph{ADD pass rate}
The $\add$ \cite{tremblay2018deep} is defined as the average Euclidean distance between the corresponding predicted ($\mathbf{p}_{\oppre}^{i}$) and reference ($\mathbf{p}_{\opref}^{i}$) vertices and centroid ($\mathbf{p}^9 \in \mathbb{R}^3$) of the object 3D bounding box:
\begin{equation}\label{eq:add_supmat}
    \add = \frac{1}{9}\sum_{i=1}^{9} ||\mathbf{p}_{\oppre}^{i} - \mathbf{p}_{\opref}^{i}||_2 \ .
\end{equation}
The $\add$ pass rate ($\add_t$) measures the percentage of frames where the $\add$ value of the prediction ($P$) is lower than a selected threshold ($t \in \mathbb{R}$):
\begin{equation}\label{eq:addt_supmat}
    \add_t = \frac{|\{P | \add \leq t\}|}{|\{P\}|}  \cdot 100\% \ .
\end{equation}
A higher $\add_t$ value for a given threshold $t$ indicates a better prediction accuracy of the object 3D bounding box.
In our experiments, we report $\add$ pass rate values for thresholds $t = $ \twocm ($\add_2$), \fivecm ($\add_5$) and \tencm ($\add_{10}$).
By definition, $\add_2 \leq \add_5 \leq \add_{10}$.

For comparison of models trained with ($\add_t^{\ratioaug}$) and without ($\add_t^{\rationoaug}$) augmentation, we use the ratio of their respective $\add$ pass rates:
\begin{equation}\label{eq:add_ratio_supmat}
    \add_t^{\ratio} = \frac{\add_t^{\ratioaug}}{\add_t^{\rationoaug}}\ .
\end{equation}
A higher $\add_t^{\ratio}$ value indicates a bigger benefit of the augmentation.


\paragraph{Rotation error} The rotation error measures the angle between the predicted ($\mathbf{R}_{\oppre}$) and reference ($\mathbf{R}_{\opref}$) rotation matrices:
\begin{equation}\label{eq:rot_error_supmat}
    E_{\rot} = \arccos{\left(\frac{\operatorname{trace}({\mathbf{R}_{\oppre}}^{-1} \mathbf{R}_{\opref}) - 1}{2}\right)} \ .
\end{equation}
A lower $E_{\rot}$ value corresponds to a better estimate of the object orientation.

\paragraph{Translation error} The translation error measures the Euclidean distance between the predicted ($\mathbf{t}_{\oppre}$) and reference ($\mathbf{t}_{\opref}$) translation vectors:
\begin{equation}\label{eq:tra_error_supmat}
    E_{\tra} = ||{\mathbf{t}_{\oppre} - \mathbf{t}_{\opref}}||_2 \ .
\end{equation}
A lower $E_{\tra}$ value indicates a better localization of the object in space.

\subsection*{A.4 Object segmentation methods for background augmentation}
\label{sec:segmentation-details}

\begin{figure}[t]
    \begin{center}
\includegraphics[width=0.98\linewidth]{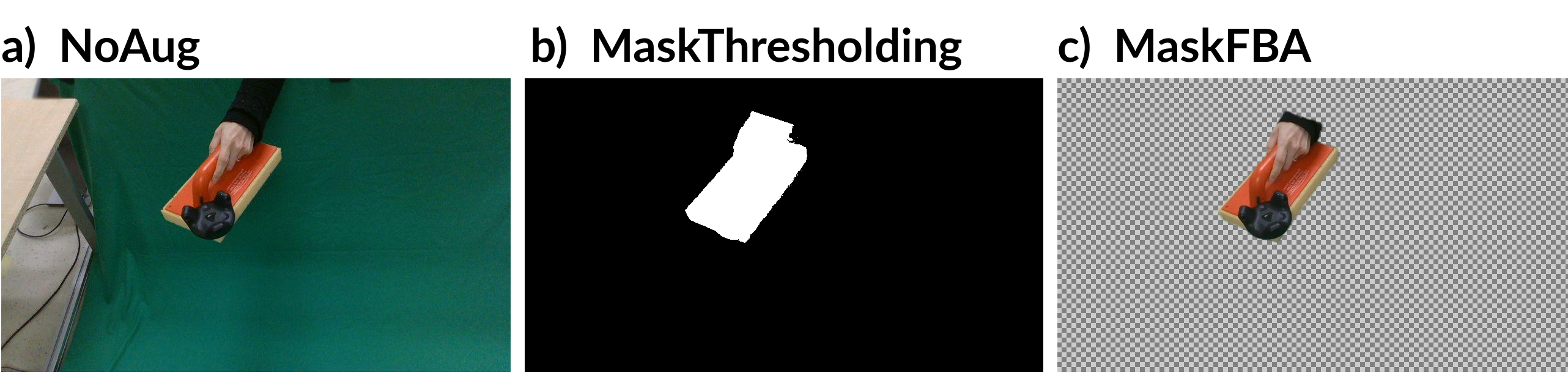}
  \end{center}
    \caption{
    Segmentation of a frame from the \train{} dataset.
    a) Original image (\noaugmentation)
    and segmentation of the tool and hand by
    b) \maskthresholding and c) \maskfba (see Sec.~A.4).
}   \label{fig:segmentations}
\end{figure}

In order to augment the image background, we need to segment the shape of the object.
Because we work with hand-held tools that are heavily occluded by the hand, we segment both the tool and the hand.
We leverage the green background in the \train{} dataset to segment the image by thresholding.
To remove the rest of the arm, we crop the segmentation mask by the convex hull of the tool 3D bounding box vertices projected into the 2D image.
The result is a binary mask of the tool and hand (\maskthresholding, see Fig.~\ref{fig:segmentations}b).
We enhance the segmentation by $F$, $B$, Alpha Matting \cite{fbamatting2020}, which estimates also the foreground opacity and color along the boundaries.
The output is an RGBA image with  opaque foreground, transparent background, and smooth boundaries between them (\maskfba, see Fig.~\ref{fig:segmentations}c).


\subsection*{A.5 Object pose estimator DOPE settings}
\label{sec:dope-details}
We estimate the \sixd object pose by the DOPE method \cite{tremblay2018deep} in each frame and concatenate the frame predictions into a complete trajectory.
Since we focus our evaluation on pose estimation in separate images, we do not post-process the individual frame predictions with any temporal or dynamic model.
Our implementation of the DOPE method was based on the referenced PyTorch implementation by \cite{tremblay2018deep}.
We trained our models on Nvidia 1080Ti and Nvidia V100 GPUs 
using the ADAM optimizer \cite{kingma2014adam}, learning rate 0.0001, and batch size 16.
To be able to train with this batch size, we downsized the input image dimensions by half to 424$\times$240 pixels, which decreased the training time without a negative impact on the accuracy (see Sec.~A.6 for the corresponding ablation study).
The ground truth belief maps we used for the DOPE training contain a 2D Gaussian with 2-pixel standard deviation and 2-pixel radius at the bounding box vertices and centroid.
In all experiments, we train on subsets of the \train{} dataset and test on subsets of the \test{} dataset.

\begin{table}[!ht]
\caption{Impact of input image resolution and batch size on DOPE \sixd object pose accuracy (see Sec.~A.6). ADD pass rates achieved by training with the original resolution (848$\times$480 pixels) and batch size 8 and with the images downsized by a factor of two (424$\times$240 pixels) and batch size 16.}
\centering
\begin{tabular}{ccccccc}
$\add_t$ & \multicolumn{3}{c}{ 848$\times$480 pixels} & \multicolumn{3}{c}{ 424$\times$240 pixels} \\
threshold $t$ & \twocm & \fivecm & \tencm & \twocm & \fivecm & \tencm \\
\hline
\gluegun & 7.4 & 49.8 & 75.6 & {\bf 9.9} & {\bf 57.8} & {\bf 79.8} \\
\groutfloat & {\bf 12.5} & {\bf 75.6} & 93.0 & 9.7 & 73.9 & {\bf 97.7} \\
\roller & {\bf 4.5} & 40.3 & 67.4 & 4.3 & {\bf 48.6} & {\bf 84.9} \\
\hline
average & {\bf 8.1} & 55.2 & 78.7 & 8.0 & {\bf 60.1} & {\bf 87.5} \\
\end{tabular}
\label{tab:resize}
\end{table}

\begin{table}[!ht]
\caption{Impact of object segmentation methods (see Sec.~A.7). ADD pass rates achieved by models trained using different object segmentation methods (see Sec.~A.4) for data augmentation.}
    \centering
\begin{tabular}{ccccccc}
$\add_t$ & \multicolumn{3}{c}{ \maskthresholding } & \multicolumn{3}{c}{ \maskfba } \\
threshold $t$ & \twocm & \fivecm & \tencm & \twocm & \fivecm & \tencm \\
\hline
\gluegun & 9.4	& 57.4	& 78.4 & {\bf 10.7} & {\bf 60.2} & {\bf 80.6} \\
\groutfloat & {\bf 10.7}	& {\bf 77.8}	& 97.1  & 10.5 & 73.0 & {\bf 97.3} \\
\roller & 4.3 &	41.2	& 81.3 & {\bf 4.3} & {\bf 50.5} & {\bf 86.3} \\
\hline
average & 8.1	& 58.8 &	85.6 & {\bf 8.5} & {\bf 61.2} & {\bf 88.1} \\
\end{tabular} 
\label{tab:masks}
\end{table}

\subsection*{A.6 Impact of image resolution and batch size}
\label{sec:resolution-details}

We explore the impact of downsizing the input images and increasing the batch size on the quality of the \sixd object pose estimation by DOPE. Table~\ref{tab:resize} presents a comparison of  ADD pass rates for the original (848$\times$480 pixels, batch size 8) and downsized (424$\times$240 pixels, batch size 16) frames. While the similarity in performance for the \twocm threshold could be attributed to a trade-off between the larger batch size and loss of detail, the increased batch size clearly improved the accuracy for the \fivecm and \tencm thresholds. Therefore, we use the 424$\times$240 pixel resolution and batch size 16 for all other experiments.


\subsection*{A.7 Comparison of object segmentation methods}
\label{sec:segmentation-method-comparison}

Table~\ref{tab:masks} compares the ADD pass rates for the \maskthresholding and \maskfba object segmentation methods (see Sec.~A.4).
Because \maskfba outperforms \maskthresholding on average for all three thresholds, we use \maskfba for object segmentation in all other experiments.

\begin{table*}
\caption{Comparison of data augmentation methods (see Sec.~A.8). ADD pass rates achieved by models trained with different data augmentation methods (see Sec.~V in the main paper).}
\centering
\begin{tabular}{cccccccccccccccc}
$\add_t$ & \multicolumn{3}{c}{ \noaugmentation } & \multicolumn{3}{c}{ \bgnoise } & \multicolumn{3}{c}{ \bgrandom } & \multicolumn{3}{c}{ \bgalternate } & \multicolumn{3}{c}{ \bgblend } \\
threshold $t$ & \twocm & \fivecm & \tencm & \twocm & \fivecm & \tencm & \twocm & \fivecm & \tencm & \twocm & \fivecm & \tencm & \twocm & \fivecm & \tencm \\
\hline
\gluegun & {\em 1.1} & {\em 17.0} & {\em 36.3} & 1.1 & 18.9 & 41.7 & 6.7 & 36.5 & 53.4 & 5.6 & 45.4 & 72.2 & {\bf 9.9} & {\bf 57.8} & {\bf 79.8} \\
\groutfloat & {\em 5.3} & {\em 45.4} & {\em 80.3} & 3.3 & 43.6 & 86.1 & 7.7 & 60.4 & 84.2 & 9.6 & 71.8 & 95.8 & {\bf 9.7} & {\bf 73.9} & {\bf 97.7} \\
\roller & {\em 3.4} & {\em 25.3} & {\em 56.6} & 3.8 & 26.3 & 52.9 & 2.9 & 39.8 & 85.5 & {\bf 6.3} & {\bf 52.2} & {\bf 85.6} & 4.3 & 48.6 & 84.9 \\
\hline
average & {\em 3.3} & {\em 29.2} & {\em 57.8} & 2.7 & 29.6 & 60.2 & 5.8 & 45.6 & 74.4 & 7.2 & 56.5 & 84.5 & {\bf 8.0} & {\bf 60.1} & {\bf 87.5} \\
\end{tabular}
\label{tab:aug_methods-details}
\end{table*}

\begin{table*}
\caption{
Generalization across camera viewpoints (see Sec.~A.9). Comparison of ADD pass rates for combinations of camera viewpoints (front camera C1 and right-hand side camera C2) between training and testing. ``\samecamera'' refers to training and testing on the same camera, ``\othercamera'' to training on one camera and testing on the other, and ``\bothcameras'' to training on both cameras.
The last row shows the average values for models trained without data augmentation (\noaugmentation).}
    \centering
\begin{tabular}{cccccccc}
$\add_t$ & Training & \multicolumn{3}{c}{ Test C1 } & \multicolumn{3}{c}{ Test C2 } \\
threshold $t$ & camera & \twocm & \fivecm & \tencm & \twocm & \fivecm & \tencm \\
\hline
\multirow{3}{*}{ \gluegun } & \samecamera & 8.7 & 61.6 & {\bf 84.6} & 5.8 & {\bf 53.3} & {\bf 82.6}  \\
& \othercamera & 0.0 & 0.2 & 1.8 &  0.1 & 2.2 & 12.0 \\
& \bothcameras & {\bf 13.7} & {\bf 63.6} & 83.4 & {\bf 6.3} & 50.6 & 75.0 \\
\hline
\multirow{3}{*}{ \groutfloat } & \samecamera & 2.4 & 50.7 & 90.1 & {\bf 6.7} & 66.0 & 96.2 \\
& \othercamera & 0.0 & 0.0 & 0.1 & 0.0 & 1.2 & 21.5 \\
& \bothcameras & {\bf 13.2} & {\bf 78.1} & {\bf 97.8} & {\bf 6.7} & {\bf 77.3} & {\bf 97.3} \\
\hline
\multirow{3}{*}{ \roller } & \samecamera & 5.9 & 56.7 & 80.9 & {\bf 0.4} & {\bf 37.2} & {\bf 89.0}  \\
& \othercamera & 0.0 & 0.1 & 1.2 & 0.0 & 0.0 & 0.0\\
& \bothcameras & {\bf 8.2} & {\bf 66.3} & {\bf 85.8} & 0.0 & 19.1 & 72.8 \\
\hline
\multirow{3}{*}{ average } & \samecamera & 5.7 & 56.3 & 85.2 & 4.3 & {\bf 52.2} & {\bf 89.3} \\
& \othercamera & 0.0 & 0.1 & 1.0 & 0.0 & 1.1 & 11.2 \\
& \bothcameras & {\bf 11.7} & {\bf 69.3} & {\bf 89.0} & {\bf 4.4} & 49.0 & 81.7 \\
\hline
\multirow{3}{*}{ \textit{\noaugmentation} } & \textit{\samecamera} & {\em 1.7} & {\em 28.7} &	{\em 66.0} & {\em 3.2}	& {\em 41.8}	& {\em 74.3}  \\
& \textit{\othercamera} & {\em 0.0} & {\em 0.0} &	{\em 0.6} & {\em 0.0}	& {\em 0.8}	& {\em 8.0} \\
& \textit{\bothcameras} & {\em 3.0}	& {\em 23.5} &	{\em 54.3} &	{\em 3.5} &	{\em 34.8} &	{\em 61.1} \\
\end{tabular} 
\label{tab:cameras-details}
\end{table*}

\begin{table*}
\caption{Generalization across left and right hand (see Sec.~A.10). Comparison of ADD pass rates for combinations of holding the tool in the left (LH) or right (RH) hand between training and testing. ``\samehand'' refers to training and testing on the same hand, ``\oppositehand'' to training on one hand and testing on the other, and ``\bothhands'' to training on both left and right hand. The last row shows the average values for models trained without data augmentation (\noaugmentation).}
    \centering
\begin{tabular}{cccccccc}
$\add_t$ & Training & \multicolumn{3}{c}{ Test LH } & \multicolumn{3}{c}{ Test RH } \\
threshold $t$ & hand & \twocm & \fivecm & \tencm & \twocm & \fivecm & \tencm \\
\hline
\multirow{3}{*}{ \gluegun } & \samehand & 5.6 & {\bf 60.9} & 77.6 & 1.8 & 25.6 & 56.9  \\
& \oppositehand & 0.7 & 8.6 & 25.0 & 1.1 & 22.8 & 54.8 \\
& \bothhands & {\bf 8.3} & 60.8 & {\bf 87.0} & {\bf 8.2} & {\bf 51.4} & {\bf 74.9} \\
\hline
\multirow{3}{*}{ \groutfloat } & \samehand & 3.2 & 60.7 & 95.3 & 3.4 & 57.1 & 93.7\\
& \oppositehand & 1.2 & 31.1 & 74.5 & 2.3 & 53.4 & 81.2 \\
& \bothhands & {\bf 7.9} & {\bf 70.8} & {\bf 97.8} & {\bf 11.0} & {\bf 83.8} & {\bf 98.3} \\
\hline
\multirow{3}{*}{ \roller } & \samehand & 5.9 & {\bf 49.9} & 83.0 & 1.6 & {\bf 37.9} & 73.4 \\
& \oppositehand & 0.8 & 23.3 & 45.1 & 0.3 & 4.9 & 22.7 \\
& \bothhands & {\bf 9.4} & {\bf 49.9} & {\bf 89.3} & {\bf 4.9} & 35.7 & {\bf 75.4} \\
\hline
\multirow{3}{*}{ average } & \samehand & 4.9 & 57.2 & 85.3 & 2.3 & 40.2 & 74.6\\
& \oppositehand & 0.9 & 21.0 & 48.2 &1.2 & 27.0 & 52.9  \\
& \bothhands & {\bf 8.5} & {\bf 60.5} & {\bf 91.4} & {\bf 8.0} & {\bf 57.0} & {\bf 82.8} \\
\hline
\multirow{3}{*}{ \textit{\noaugmentation} } & \textit{\samehand} & {\em 1.9} & {\em 29.4} & {\em 57.1} & {\em 1.5} & {\em 22.5} & {\em 51.9} \\
& \textit{\oppositehand} & {\em 0.9}	& {\em 13.7} &	{\em 35.0} & {\em 0.8}	& {\em 18.4}	& {\em 47.6} \\
& \textit{\bothhands} & {\em 4.0}	& {\em 31.3}	& {\em 58.4} &	{\em 2.0}	& {\em 25.6}	& {\em 52.6} \\
\end{tabular} 
\label{tab:hands-details}
\end{table*}

\begin{table*}
\caption{
Generalization across demonstrators (see Sec.~A.11). Comparison of ADD pass rates for various combinations of the four demonstrators (S1-S4) between training and testing. ``\allsubjects'' refers to training one model for all subjects, ``\differentsubject'' to training on three subjects and testing on the remaining one, and ``\samesubject'' to training and testing on the same subject. 
The ADD pass rates are averaged across all test subjects (\ie S1-S4). The last row shows the average for models trained without data augmentation (\noaugmentation).}
\centering
\begin{tabular}{cccccccccc}
$\add_t$ & \multicolumn{3}{c}{ \allsubjects } & \multicolumn{3}{c}{ \differentsubject } & \multicolumn{3}{c}{ \samesubject }\\
threshold $t$ & \twocm & \fivecm & \tencm & \twocm & \fivecm & \tencm & \twocm & \fivecm & \tencm \\
\hline
\multirow{1}{*}{ \gluegun }  & {\bf 10.1} & {\bf 57.2} & {\bf 80.1} & 6.2 & 52.6 & 78.2 & 3.4 & 32.8 & 69.4 \\
\multirow{1}{*}{ \groutfloat } & {\bf 9.5} & {\bf 77.2} & {\bf 97.5} & 6.0 & 64.1 & 95.3 & 1.8 & 38.8 & 81.6\\
\multirow{1}{*}{ \roller } & 3.9 & {\bf 41.8} & {\bf 78.1} & {\bf 4.5} & 39.1 & 77.3 & 0.7 & 21.8 & 53.8\\
\hline
average  & {\bf 7.9} & {\bf 58.8} & {\bf 85.2} & 5.6 & 52.0 & 83.6 & 2.0 & 31.1 & 68.3\\
\hline
\textit{\noaugmentation} & {\em 3.1} & {\em 28.9} & {\em 57.7} & {\em 1.9} & {\em 22.3} & {\em 52.3} & {\em 1.2} & {\em 18.7} & {\em 51.6} \\
\end{tabular} 
\label{tab:subjects-details}
\end{table*}

\subsection*{A.8 Benefits of data augmentation}
\label{sec:dataAugmentation_res-details}



Table~\ref{tab:aug_methods-details} shows \twocm, \fivecm, and \tencm ADD pass rates
for the DOPE estimator using different background augmentation methods from Sec.~IV-B in the main paper.
For all three thresholds, the best average ADD pass rates were achieved by the \bgblend augmentation.
$\add_{2}^{\ratio} = 2.4$, $\add_{5}^{\ratio} = 2.1$, and $\add_{10}^{\ratio} = 1.5$ values for \bgblend indicate a big improvement in accuracy with respect to training without augmentation (\noaugmentation).


\subsection*{A.9 Generalization across camera viewpoints}
\label{sec:cameras-details}
Table~\ref{tab:cameras-details} shows the \twocm, \fivecm, and \tencm ADD pass rates
for various combinations of the camera viewpoints between training and testing (see Sec.~V in the main paper for a detailed description of the scenarios).

\subsection*{A.10 Generalization across left/right hand}
\label{sec:hands-details}
Table~\ref{tab:hands-details} shows the \twocm, \fivecm, and \tencm ADD pass rates 
for various combinations of the left/right hand between training and testing (see Sec.~V in the main paper for a detailed description of the scenarios).

\subsection*{A.11 Generalization across demonstrators}
\label{sec:subjects-details}
Table~\ref{tab:subjects-details} shows the \twocm, \fivecm, and \tencm ADD pass rates 
for various combinations of the demonstrators between training and testing (see Sec.~V in the main paper for a detailed description of the scenarios).

\subsection*{A.12 Robustness to clutter}
\label{sec:clutter-details}
Table~\ref{tab:clutter-details} shows the \twocm, \fivecm, and \tencm ADD pass rates 
for the \sixd object pose estimator DOPE 
for the presence/absence of clutter in the test environment (see Sec.~V in the main paper for a detailed description of the scenarios).

\subsection*{A.13 Performance on different tools and tasks}
\label{sec:tasks-details}
Table~\ref{tab:tasks-details} shows the \twocm, \fivecm, and \tencm ADD pass rates as well as the rotation and translation errors (see Sec.~A.3) for all tools and tasks in the \test{} dataset.




\subsection*{A.14 Comparison of DOPE and CosyPose results}
\label{sec:cosy-pose}

\begin{figure}[t]
\centering
    \centering
    \subfloat[\powerdrill\label{fig:drill}]{
    \includegraphics[width=0.4655\linewidth]{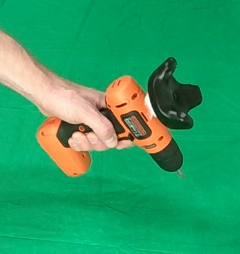} }
    \subfloat[3D model + aligned meshes\label{fig:drill_mesh}]{
    \includegraphics[width=0.47\linewidth]{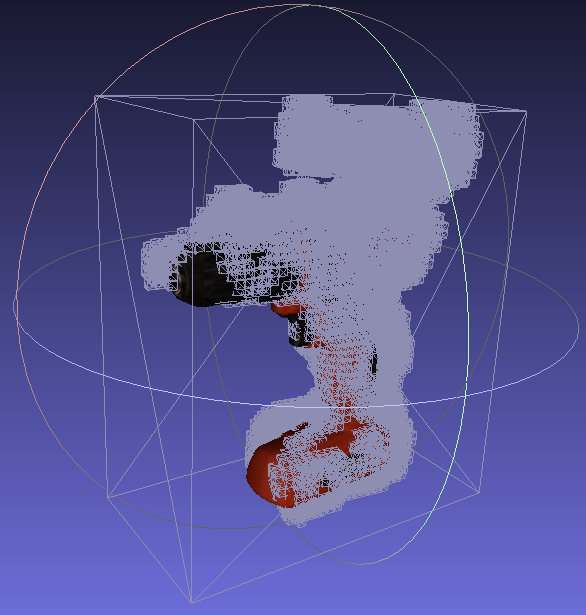} }
    \hfill
    \caption{Alignment of the \powerdrill tool meshes. a) The \powerdrill in the \train{} dataset. b) The mesh of the 3D model from the YCB Object dataset~\protect\cite{calli15ycb} aligned to the tool tracing mesh using ICP in MeshLab. The geometric transformation describes the difference between the 3D model mesh origin and the HTC Vive tracker origin.}
    \label{fig:drill_mesh_alignment}
\end{figure}

The \powerdrill has a 3D model available in the YCB Object dataset~\cite{calli15ycb}, which enables comparison of model-free and model-based object pose estimation methods on this tool.
Here we compare the performance of model-free estimator DOPE~\cite{tremblay2018deep} and model-based estimator CosyPose~\cite{labbe2020cosypose}.
While CosyPose had been trained extensively on rendered images of 3D models of the objects from the YCB Object dataset, DOPE was trained on short video sequences of the hand-held \powerdrill from the \train{} dataset (see Fig.~\ref{fig:drill}).
Below we compare the detection rates and the rotation and translation errors of DOPE and CosyPose on \powerdrill in the \test{} dataset.

\begin{table}[t]
\caption{Robustness to clutter (see Sec.~A.12). Comparison of ADD pass rates for \gluegun task frame tested on a table with only the gluing frame (\noclutter) and with a clutter of other objects around the frame (\clutter). The bottom row shows results for a model trained without data augmentation (\noaugmentation).}
\centering
\begin{tabular}{ccccccc}
$\add_t$ & \multicolumn{3}{c}{ \noclutter } & \multicolumn{3}{c}{ \clutter } \\
threshold $t$ & \twocm & \fivecm & \tencm & \twocm & \fivecm & \tencm \\
\hline
{ \gluegun (frame) } & {\bf 8.6} & {\bf 61.8} & {\bf 90.1} & {\bf 11.0} & {\bf 61.5} & {\bf 83.7} \\
\hline
{ \textit{\noaugmentation}} & {\em 1.7} & {\em 22.8} & {\em 47.7} & {\em 0.3} & {\em 4.9} & {\em 19.8} \\
\end{tabular}
\label{tab:clutter-details}
\end{table}

\begin{table}[t]
\caption{Performance of the 6D object pose estimator DOPE on different tools and manipulation tasks (see Sec.~A.13). \twocm, \fivecm, and \tencm ADD pass rate accuracy ($\add_t$) and average rotation ($E_{\rot}$) and translation ($E_{\tra}$) errors for different tools and tasks.
Invalid detections were excluded from the computation of average $E_{\rot}$ and $E_{\tra}$.}
\centering
\begin{tabular}{ccccccc}
\multirow{2}{*}{Tool} & \multirow{2}{*}{Task} & \multicolumn{3}{c}{ $\add_t$ (\%)} & $E_{\rot}$ & $E_{\tra}$ \\
{ } & { } & \twocm & \fivecm & \tencm & (deg) & (cm)\\
\hline
\multirow{4}{*}{\gluegun} & frame & 8.0 & 53.3 & 77.1 & 11.8 & 5.0\\
& densewave & {\bf 10.6} & 61.9 & 88.6 & 5.0 & 3.6\\
& sparsewave & 8.3 & 66.0 & 91.0 & 5.0 & 3.4\\
\cline{2-7}
& average & 9.0 & 60.4 & 85.6 & 7.3 & 4.0 \\
\hline
\multirow{3}{*}{\groutfloat} & round & 9.2 & 74.4 & {\bf 98.7} & {\bf 3.9} & 2.7\\
& sweep & 9.3 & {\bf 82.7} & 98.1 & 4.3 & {\bf 2.2}\\
\cline{2-7}
& average & 9.3 & 78.6 & 98.4 & 4.1 & 2.5 \\
\hline
\roller & press & 4.3 & 50.5 & 86.3 & 8.7 & 3.7\\
\gluegunII & lshape & 0.0 & 9.0 & 41.9 & 38.5 & 9.9 \\
\gluegunIII & lshape & 0.1 & 4.7 & 30.0 & 40.3 & 10.2 \\
\gluegunIIIdtp & lshape & 1.2 & 23.4 & 52.6 & 20.9 & 8.4 \\
\heatgun & heating & 0.0 & 13.2 & 56.3 & 14.3 & 7.0 \\
\powerdrill & down & 5.6 & 59.8 & 87.0 & 8.0 & 3.8 \\
\solder & soldering & 0.5 & 12.8 & 41.4 & 35.6 & 9.0 \\
\hline
average & - & 3.3 & 34.7 & 64.4 & 19.8 & 6.5 \\
\end{tabular} 
\label{tab:tasks-details}
\end{table}

Using 80\% confidence threshold, CosyPose detected the \powerdrill in 34\% of the test frames.
Additional wrong objects were detected in less than 1\% of the frames, but the wrong detections had significantly lower confidence than the correct ones and there were no cases where only wrong objects were detected.
No object was detected in the remaining 66\% of the frames.

To estimate the rotation and translation errors, we needed to convert our ground truth annotations to have the same reference coordinate system as the 3D model.
Therefore, we aligned the 3D model mesh from the YCB Object dataset with our tracing mesh using ICP in  MeshLab (see~Fig.~\ref{fig:drill_mesh}).
The average rotation and translation errors of CosyPose were $E_{\rot} = 43.6^\circ$ and $E_{\tra} = 4.9$\,cm, respectively.


However, the comparison with the DOPE results (see Table~\ref{tab:cosypose}) is not straightforward.
Most importantly, the performace of CosyPose may be affected by the presence of the HTC Vive tracker in the \test{} dataset.
In the case of DOPE, the tracker was present both in training and testing, whereas CosyPose was trained using a 3D model of the tool without the tracker.
In addition, while DOPE was trained on a single object (\powerdrill), CosyPose was trained on a set of multiple objects (YCB Object dataset), leading to possible false positives.
Also, the DOPE results were filtered to exclude detections farther than one meter from the reference pose, but this affected less than one percent of frames. 


\begin{table}[t]
    \centering
    \caption{Comparison of model-free estimator DOPE~\protect\cite{tremblay2018deep} and model-based estimator CosyPose~\protect\cite{labbe2020cosypose} on the \powerdrill tool. The percentage of frames where the tool was detected (Detections) and average rotation ($E_{\rot}$) and translation ($E_{\tra}$) errors.}
    \begin{tabular}{cccc}
        \sixd object pose & Detections & $E_{\rot}$ & $E_{\tra}$ \\
        estimation method & (\% frames) & (deg) & (cm)\\
        \hline
        DOPE & {\bf 99.3\%} & {\bf 8.0} & {\bf 3.8} \\
        CosyPose & 34.0\% & 43.6 & 4.9 \\
    \end{tabular}
    \label{tab:cosypose}
\end{table}

\subsection*{A.15 Robustness to tracker position}
\label{sec:tracker-position}

\begin{figure}[t]
    \centering
    \subfloat[\gluegunIII\label{fig:dif_trackerSetup_III}]{
    \includegraphics[width=0.47\linewidth]{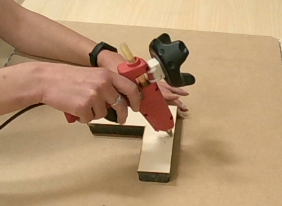} }
    \subfloat[\gluegunIIIdtp\label{fig:dif_trackerSetup_IIIdtp}]{
    \includegraphics[width=0.47\linewidth]{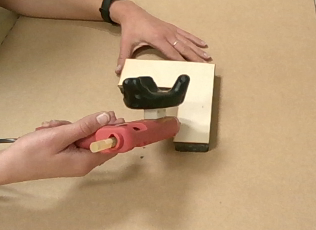} }
    \caption{The same object with different HTC Vive tracker positions: a) \gluegunIII has the tracker mounted on the top; b) \gluegunIIIdtp has the tracker mounted on the left side.}
    \label{fig:dif_trackerSetup}
\end{figure}

To evaluate the impact of the HTC Vive tracker position on the tool on the accuracy of the \sixd object pose estimator, we have recorded the same object with two different tracker positions: \gluegunIII has the tracker mounted on the top, while \gluegunIIIdtp has the tracker mounted on its left side (see Fig.~\ref{fig:dif_trackerSetup}).
We have trained and evaluated the \sixd object pose estimator DOPE using four different configurations:
a) training and testing on \gluegunIII,
b) training and testing on \gluegunIIIdtp (these two configurations are reported also in the main paper),
c) training on \gluegunIII and testing on \gluegunIIIdtp, and
d) training on \gluegunIIIdtp and testing on \gluegunIII.
Table~\ref{tab:dif_tracker} shows the resulting $\add_5$ accuracy and average translation and rotation errors.

In the \gluegunIII to \gluegunIII and \gluegunIIIdtp to \gluegunIIIdtp configurations, the pose estimator performed better on \gluegunIIIdtp than on \gluegunIII.
This may be related to smaller occlusions of the tool when the tracker is mounted on its left side rather than on the top, considering that the side camera (C2) is on the right-hand side in out setup.
However, in the \gluegunIII to \gluegunIIIdtp and \gluegunIIIdtp to \gluegunIII configurations, where the position of the tracker changed between training and testing, the estimator was not able to correctly predict the pose of the tool.

\begin{table}[t]
\centering
\caption{Robustness to tracker position (see Sec.~A.15). 5~cm ADD pass rate accuracy ($\add_5$) and average rotation ($E_{\rot}$) and translation ($E_{\tra}$) errors for different combinations of training and testing on \gluegunIII (tracker mounted on the top) and \gluegunIIIdtp (tracker mounted on the left).}
\label{tab:dif_tracker}
\begin{tabular}{ccccc}
{Training} & {Testing} & \multicolumn{1}{c}{ $\add_5$} & $E_{\rot}$ & $E_{\tra}$ \\
{tool} & {tool} & (\%) & (deg) & (cm)\\
\hline
\gluegunIII & \gluegunIII & 4.7 & 40.3 & 10.2 \\
\gluegunIIIdtp & \gluegunIIIdtp & {\bf 23.4} & {\bf 20.9} & {\bf 8.4} \\
\hline
\gluegunIII & \gluegunIIIdtp & 0.0 & 148.6 & 18.3 \\
\gluegunIIIdtp & \gluegunIII & 0.0 & 148.1 & 19.8 \\
\end{tabular} 
\end{table}

These experiments indicate that the selected tracker position can affect the \sixd object pose estimator performance and that a transfer between different tracker positions is a challenging problem.
For a real application, the HTC Vive tracker could be replaced by a smaller or concealed tracking device, e.g.~based on an inertial measuring unit.
However, this engineering task is beyond the scope of this paper, as the main goal of the \dataset dataset is benchmarking \sixd object pose estimation methods on hand-held tool manipulation tasks rather than deployment to the end-user.

\section*{B. Dataset Documentation and Intended Uses}
\label{sec:dataset_documentation}

\paragraph*{Dataset documentation} The \dataset dataset documentation, metadata and download instructions are available at: \url{http://imitrob.ciirc.cvut.cz/imitrobdataset.php}

\paragraph*{Supplementary code} The GitHub repository for the supplementary code (including example usage of the DOPE~\cite{tremblay2018deep} method) is at: \url{https://github.com/imitrob/imitrob_dataset_code}

\paragraph*{Intended uses} The dataset is primarily intended for benchmarking \sixd pose estimation methods in manipulation tasks with hand-held objects and evaluating their ability to generalize with respect to various conditions. It can be also used to evaluate the effect of different data augmentation methods. Another usage is the methodology for data acquisition and \sixd pose estimator training for new tools and tasks and a guideline for collecting more extensive datasets and benchmarking \sixd object pose estimators on various tasks with hand-held tools, \eg in imitation learning, grasping, virtual or augmented reality, etc. In general, we hope that the presented dataset will trigger further development of 6D object pose estimation methods and their usage in various industrial tasks based on the required accuracy.

\paragraph*{Author statement} We bear all responsibility in case of violation of right in using our dataset or code. We confirm that we used all the existing assets in accordance to their license.

\section*{C. Hosting, Licensing, and Maintainance Plan}
\label{sec:hosting_licensing}

\paragraph*{Hosting} The \dataset dataset is hosted on our in-house servers, which are managed by our dedicated IT department. The dataset and source code are publicly available. The dataset website (\url{http://imitrob.ciirc.cvut.cz/imitrobdataset.php}) describes the dataset and provides download links. The source code is hosted on \url{https://github.com/imitrob/imitrob_dataset_code}.  

\paragraph*{Maintainance} The authors will provide important bug fixes to the community as commits to the GitHub repository. The dataset webpage will summarize changes to the code and the dataset. In the unlikely case that our in-house data center stops operating, we will migrate the dataset to another hosting and announce the new links in the GitHub repository.

\paragraph*{Licensing} The provided dataset and supplementary code are copyrighted by us and published under the CC~BY-NC-SA~4.0 license (\url{https://creativecommons.org/licenses/by-nc-sa/4.0/}). 
To use the code or the dataset, the original work has to be attributed, as specified by the authors on the dataset or code repository websites.

\paragraph*{Contributions} Contributions to the dataset and supplementary code are welcome and contributors should contact the authors.

\paragraph*{Contact} The contact e-mail address of the manager of the dataset: karla.stepanova@cvut.cz.


\section*{D. Datasheet for Dataset Imitrob}

\makeatletter
\renewcommand{\@IEEEsectpunct}{\smallskip\hfill\\}
\makeatother

Questions from the Datasheets for Datasets (\url{https://arxiv.org/abs/1803.09010}) paper, v7.
\subsection*{D.1 Motivation}

\subsubsection*{For what purpose was the dataset created?}
The \dataset dataset was created with the aim to enable imitation learning of manipulation tasks purely from visual observations. This includes the ability to recognize \sixd pose of the hand-held objects. Current methods are typically trained and tested in different conditions than this kind of tasks, so it is very difficult to estimate how they will perform in manipulation tasks with hand-held tools. As expected, the tested methods showed quite low accuracy in the case of the mainipulation with hand-held tools, especially when generalization to new users, camera viewpoints, or tasks was needed. This motivated the creation of a new dataset, which would enable benchmarking of these methods.

\bigskip

\subsubsection*{Who created the dataset (e.g., which team, research group) and on behalf of which entity (e.g., company, institution, organization)?} 

The dataset was created by Jiri Sedlar, Karla Stepanova, Radoslav Skoviera, Gabriela Sejnova, Jan K.~Behrens, and Josef Sivic within CIIRC CTU in Prague (Imitation learning centre \url{http://imitrob.ciirc.cvut.cz}) in collaboration with Matus Tuna from Comenius University in Bratislava and Robert Babuska from TU Delft.

\bigskip

\subsubsection*{Who funded the creation of the dataset?} 

Jiri Sedlar and Josef Sivic were supported by the European Regional Development Fund under the project IMPACT (reg.~no.~CZ.02.1.01/0.0/0.0/15\_003/0000468)
and the EU Horizon Europe Programme under the project AGIMUS (reg.~no.~101070165). Matus Tuna was supported by project VEGA 1/0796/18. Karla Stepanova, Radoslav Skoviera, and Gabriela Sejnova were supported by the Technological Agency of CR under the grant Collaborative workspace of the future (reg.~no.~FV40319). Gabriela Sejnova was supported by CTU Student Grant Agency (reg.~no.~SGS21/184/OHK3/3T/37). Radoslav Skoviera, Jan Kristof Behrens, and Robert Babuska were supported by the European Regional Development Fund under the project Robotics for Industry 4.0 (reg.~no.~CZ.02.1.01/0.0/0.0/15\_003/0000470).

\subsection*{D.2 Composition}

\subsubsection*{What do the instances that comprise the dataset represent (e.g., documents, photos, people, countries)?}

The dataset consists of RGB-D images extracted from 352 video sequences, accompanied by 6D annotation. The videos capture simple manipulation tasks with 9 hand-held tools (\gluegun, \groutfloat, \roller, \gluegunII, \gluegunIII, \gluegunIIIdtp, \heatgun, \powerdrill, and \solder) such as applying glue along a given trajectory, polishing a surface, or flattening a cloth.

\bigskip

\subsubsection*{How many instances are there in total (of each type, if appropriate)?}

The \dataset dataset contains images extracted from 352 video sequences (208 in the \test{} dataset and 144 in the \train{} dataset) of hand-held tool manipulations. The \test{} component of the dataset contains 100\,332 images and the \train component contains 83\,778 images.

\bigskip

\subsubsection*{Does the dataset contain all possible instances or is it a sample (not necessarily random) of instances from a larger set?}

The dataset contains all the possible instances.

\bigskip

\subsubsection*{What data does each instance consist of?}

Each video frame contains the following data:
\begin{itemize}
\item 6D pose of the recorded tool {\fontfamily{cmtt}\selectfont (6DOF/*.json)}
\item 2D image coordinates of the tool 3D bounding box vertices and centroid {\fontfamily{cmtt}\selectfont (BBox/*.json)}
\item depth image {\fontfamily{cmtt}\selectfont (Depth/*.png)}
\item RGB image {\fontfamily{cmtt}\selectfont (Image/*.png)}
\end{itemize}
In addition, each frame of the \train{} dataset also contains:
\begin{itemize}
\item binary mask of the segmented tool and hand {\fontfamily{cmtt}\selectfont (Mask\_thresholding/*.png)}
\item RGB image with the segmented tool and hand opaque and the background transparent {\fontfamily{cmtt}\selectfont (Mask/*.png)}
\end{itemize}
Each video sequence in the \dataset dataset contains:
\begin{itemize}
\item 3D coordinates of the tool bounding box vertices and centroid with respect to the HTC Vive Tracker ({\fontfamily{cmtt}\selectfont BB\_in\_tracker}) and intrinsic camera matrices for cameras C1 ({\fontfamily{cmtt}\selectfont K\_C1}) and C2 ({\fontfamily{cmtt}\selectfont K\_C2}) {\fontfamily{cmtt}\selectfont (parameters.json)}
\end{itemize}
The training/test component, tool, task, subject, camera, left/right hand or presence/absence of clutter are identified in the name of the video sequence folder.

\bigskip

\subsubsection*{Is there a label or target associated with each instance?}

Yes, each image is annotated with the \sixd pose of the tool as well as the video sequence labels, including the identifier of the training/test component, tool, task, subject, camera viewpoint, left/right hand, or presence/absence of clutter.

\bigskip

\subsubsection*{Is any information missing from individual instances?}

The \sixd pose for individual data frames was interpolated. When the time difference between consecutive HTC Vive frames was longer than 100 ms, the corresponding camera images were discarded to ensure sufficient accuracy of the ground truth data. Otherwise no information is missing and the data is complete.

\bigskip

\subsubsection*{Are relationships between individual instances made explicit (e.g., users’ movie ratings, social network links)?}

Yes, the relationships are fully identified by the video sequence labels (see above) and the position of the frame in the sequence.

\bigskip

\subsubsection*{Are there recommended data splits (e.g., training, development/validation, testing)?}

We explicitly state the data splits used for training and testing of the \sixd pose estimator. The training set is an (augmented) subset of the \train{} dataset, and the test set is a subset of the \test{} dataset.

\bigskip

\subsubsection*{Are there any errors, sources of noise, or redundancies in the dataset?}

Sources of noise include the calibration of the cameras and the HTC Vive controllers and the synchronization between the HTC Vive and the cameras (the HTC Vive data were interpolated to the closest camera frame and if the distance between two consecutive frames was longer than 100 ms, the corresponding camera images were discarded to ensure sufficiently accurate ground truth data).

\bigskip

\subsubsection*{Is the dataset self-contained, or does it link to or otherwise rely on external resources (e.g., websites, tweets, other datasets)?}

Both the dataset and the supplementary code are self-contained.

\bigskip

\subsubsection*{Does the dataset contain data that might be considered confidential (e.g., data that is protected by legal privilege or by doctor-patient confidentiality, data that includes the content of individuals’ non-public communications)?}

N/A.

\bigskip

\subsubsection*{Does the dataset contain data that, if viewed directly, might be offensive, insulting, threatening, or might otherwise cause anxiety?}

N/A.

\bigskip

\subsubsection*{Does the dataset relate to people? }

N/A.

\bigskip

\subsubsection*{Does the dataset identify any subpopulations (e.g., by age, gender)?}

N/A.

\bigskip

\subsubsection*{Is it possible to identify individuals (i.e., one or more natural persons), either directly or indirectly (i.e., in combination with other data) from the dataset?}

N/A.

\bigskip

\subsubsection*{Does the dataset contain data that might be considered sensitive in any way (e.g., data that reveals racial or ethnic origins, sexual orientations, religious beliefs, political opinions or union memberships, or locations; financial or health data; biometric or genetic data; forms of government identification, such as social security numbers; criminal history)?}

N/A.

\bigskip

\subsubsection*{Any other comments?}

N/A.

\subsection*{D.3 Collection process}

\subsubsection*{How was the data associated with each instance acquired?}

The directly observable data (RGB-D images) were synchronized with observable HTC Vive data. The parameters of each video sequence setup (such as the tool, task, subject, camera viewpoint, left/right hand, or presence/absence of clutter) were manually annotated and associated with the corresponding data.

\bigskip

\subsubsection*{What mechanisms or procedures were used to collect the data (e.g., hardware apparatus or sensor, manual human curation, software program, software API)?}

The visual part of the dataset was collected by two RGB-D cameras, specifically Intel RealSense D-435. The resolution of both RGB and depth images was set to 848x480 and they were recorded at 60 FPS. The 6D pose information was recorded using HTC Vive VR system in standard configuration. An HTC Vive tracker was mounted to the tools to acquire their pose. The cameras and HTC Vive system were calibrated towards a common coordinate system. The calibration of the camera and HTC Vive was validated by the average distances of associated points using a checkerboard pattern.
The whole acquisition system was implemented via the Robot Operating System, using Python as the main programming language.

\bigskip

\subsubsection*{If the dataset is a sample from a larger set, what was the sampling strategy (e.g., deterministic, probabilistic with specific sampling probabilities)?}

N/A.

\bigskip

\subsubsection*{Who was involved in the data collection process (e.g., students, crowdworkers, contractors) and how were they compensated (e.g., how much were crowdworkers paid)?}

Only the authors were involved in the collection process.

\bigskip

\subsubsection*{Over what timeframe was the data collected?}

The data was collected in January and February 2021 (\gluegun, \groutfloat, and \roller) and in March 2023 (\gluegunII, \gluegunIII, \gluegunIIIdtp, \heatgun, \powerdrill, \solder).

\bigskip

\subsubsection*{Were any ethical review processes conducted (e.g., by an institutional review board)?}

N/A.

\bigskip

\subsubsection*{Does the dataset relate to people?}

N/A.

\subsection*{D.4 Preprocessing/cleaning/labeling}

\subsubsection*{Was any preprocessing/cleaning/labeling of the data done (e.g., discretization or bucketing, tokenization, part-of-speech tagging, SIFT feature extraction, removal of instances, processing of missing values)?}

The data were originally recorded as ROS bag files, from which the individual data instances were extracted, synchronized, interpolated, and saved to separate folders.
For the \train{} dataset, the masks were created by automatic segmentation of the RGB images.

\bigskip

\subsubsection*{Was the “raw” data saved in addition to the preprocessed/cleaned/labeled data (e.g., to support unanticipated future uses)?}

The original bag files are saved on our internal data storage, but are too big to be easily shareable.

\bigskip

\subsubsection*{Is the software used to preprocess/clean/label the instances available?}

No, we don't provide the raw data and thus neither the code to process it. Dataset manipulation tools (for the already preprocessed and labeled data) are available on the supplementary code GitHub page: \url{https://github.com/imitrob/imitrob_dataset_code}.

\subsection*{D.5 Uses}

\subsubsection*{Has the dataset been used for any tasks already?}

This is the first use of the dataset.

\bigskip

\subsubsection*{Is there a repository that links to any or all papers or systems that use the dataset?}

There are no papers that use our dataset, yet. Future uses will be added to the dataset/code website.

\bigskip

\subsubsection*{What (other) tasks could the dataset be used for?}
The dataset is primarily intended for benchmarking \sixd pose estimation methods in manipulation tasks with hand-held objects and evaluating their ability to generalize with respect to various conditions. It can be also used to evaluate the effect of different data augmentation methods. Another usage is the methodology for data acquisition and \sixd pose estimator training for new tools and tasks and a guideline for collecting more extensive datasets and benchmarking \sixd object pose estimators on various tasks with hand-held tools, \eg in imitation learning, grasping, virtual or augmented reality, etc. In general, we hope that the presented dataset will trigger further development of 6D object pose estimation methods and their usage in various industrial tasks based on the required accuracy.

\bigskip

\subsubsection*{Is there anything about the composition of the dataset or the way it was collected and preprocessed/cleaned/labeled that might impact future uses?}

N/A.

\bigskip

\subsubsection*{Are there tasks for which the dataset should not be used?}

N/A.

\subsection*{D.6 Distribution}

\subsubsection*{Will the dataset be distributed to third parties outside of the entity (e.g., company, institution, organization) on behalf of which the dataset was created?} 

N/A.

\bigskip

\subsubsection*{How will the dataset will be distributed (e.g., tarball on website, API, GitHub)?}

The dataset is available on the dataset website: \url{http://imitrob.ciirc.cvut.cz/imitrobdataset.php}

\bigskip

\subsubsection*{When will the dataset be distributed?}

In 2023.

\bigskip

\subsubsection*{Will the dataset be distributed under a copyright or other intellectual property (IP) license, and/or under applicable terms of use (ToU)?}

The newly provided datasets and benchmarks are copyrighted by us and published under the CC BY-NC-SA 4.0 license (\url{https://creativecommons.org/licenses/by-nc-sa/4.0/}).

\bigskip

\subsubsection*{Have any third parties imposed IP-based or other restrictions on the data associated with the instances?}

N/A.

\bigskip

\subsubsection*{Do any export controls or other regulatory restrictions apply to the dataset or to individual instances?}

N/A.

\subsection*{D.7 Maintenance}

\subsubsection*{Who is supporting/hosting/maintaining the dataset?}

Karla Stepanova at CIIRC CTU in Prague.

\bigskip

\subsubsection*{How can the owner/curator/manager of the dataset be contacted (e.g., email address)?}

Contact e-mail address: karla.stepanova@cvut.cz

\bigskip

\subsubsection*{Is there an erratum?}

Any updates to the code will be visible as commits in the GitHub repository.
The dataset website will summarize all changes to the code and the dataset.

\bigskip

\subsubsection*{Will the dataset be updated (e.g., to correct labeling errors, add new instances, delete instances)?}

Any updates to the code will be visible as commits in the GitHub repository.
The dataset website will summarize all changes to the code and the dataset.

\bigskip

\subsubsection*{If the dataset relates to people, are there applicable limits on the retention of the data associated with the instances (e.g., were individuals in question told that their data would be retained for a fixed period of time and then deleted)?}

N/A

\bigskip

\subsubsection*{Will older dataset versions continue to be supported/hosted/maintained?}

N/A.

\bigskip

\subsubsection*{If others want to extend/augment/build on/contribute to the dataset, is there a mechanism for them to do so?}

Yes, contributions to the dataset are welcome. Please get in touch with the maintainer of the dataset via e-mail (see above).


